%% file: main.tex
\pdfoutput=1
\documentclass[sigconf,nonacm]{acmart}
\usepackage{pgf-pie}
\usepackage{pgfplots}
\usepackage{tikz}
\pgfplotsset{compat=1.18} 
\usetikzlibrary{patterns}
\usepackage{caption}
\usepackage{subcaption}
\usepackage{makecell}
\setcellgapes{1pt}
\usepackage{listings}
\usepackage{pifont}

\newcommand{\xmark}{\ding{55}}

\definecolor{absolute}{HTML}{29855E}
\definecolor{absolute_rand}{HTML}{29855E}
\definecolor{relative}{HTML}{85602A}
\definecolor{relative_rand}{HTML}{85602A}

\definecolor{absolute_davinci}{HTML}{853848}
\definecolor{relative_davinci}{HTML}{4B827B}
\definecolor{absolute_davinci_random}{HTML}{853848}
\definecolor{relative_davinci_random}{HTML}{4B827B}

\definecolor{absolute_alpaca}{HTML}{29855E}

\definecolor{absolute_falcon}{HTML}{806E61}
\definecolor{relative_falcon}{HTML}{AFCCBE}

\definecolor{absolute_openllama}{HTML}{853848}
\definecolor{relative_openllama}{HTML}{61853E}

\definecolor{absolute_redpajama_3}{HTML}{6539DB}
\definecolor{relative_redpajama_3}{HTML}{DBB044}

\definecolor{absolute_redpajama_7}{HTML}{556A85}
\definecolor{relative_redpajama_7}{HTML}{857D4E}

\newcommand{\alpaca}{\texttt{alapaca-7B}}
\newcommand{\dvc}{\texttt{text-davinci-003}}
\newcommand{\openllama}{\texttt{open-llama-7B}}
\newcommand{\rpsmall}{\texttt{red-pajama-3B}}
\newcommand{\rpbig}{\texttt{red-pajama-7B}}
\newcommand{\falcon}{\texttt{falcon-7B}}

\begin{document}

\title{Temporal Blind Spots in Large Language Models}

\author{Jonas Wallat}
\affiliation{%
  \institution{L3S Research Center}
  \streetaddress{Campus 1.4}
  \city{Hannover}
  \state{Germany}
}
\email{jonas.wallat@l3s.de}

\author{Adam Jatowt}
\affiliation{%
  \institution{University of Innsbruck}
  \streetaddress{Campus 1.4}
  \city{Innsbruck}
  \state{Austria}
}
\email{adam.jatowt@uibk.ac.at}

\author{Avishek Anand}
\affiliation{%
  \institution{Delft University of Technology}
  \streetaddress{Campus 1.4}
  \city{Delft}
  \state{The Netherlands}
}
\email{avishek.anand@tudelft.nl}

\renewcommand{\shortauthors}{Jonas Wallat, Adam Jatowt, \& Avishek Anand}

\settopmatter{printacmref=false, printccs=false, printfolios=false}

\input{abstract}
\maketitle

\input{introduction}

\input{related-work}

\input{study}

\input{results}

\input{temp-errors}

\input{conclusion}

\begin{acks}
This research was partially funded by the Federal Ministry of Education and Research (BMBF), Germany under the project LeibnizKILabor with grant No. 01DD20003 and Cubra with grant No. 13N16052. It is also partially supported by the EU – Horizon 2020 Program, Grant Agreement n.871042, “SoBigData++”.
\end{acks}

\newpage

\bibliographystyle{abbrv}
\bibliography{references}

\newpage

\appendix
\input{appendix}

\end{document}

%% file: abstract.tex
\begin{abstract}
Large language models (LLMs) have recently gained significant attention due to their unparalleled ability to perform various natural language processing tasks. These models, benefiting from their advanced natural language understanding capabilities, have demonstrated impressive zero-shot performance. However, the pre-training data utilized in LLMs is often confined to a specific corpus, resulting in inherent freshness and temporal scope limitations. Consequently, this raises concerns regarding the effectiveness of LLMs for tasks involving temporal intents.
In this study, we aim to investigate the underlying limitations of general-purpose LLMs when deployed for tasks that require a temporal understanding. 
We pay particular attention to handling factual temporal knowledge through three popular temporal QA datasets. Specifically, we observe low performance on detailed questions about the past and, surprisingly, for rather new information. 
In manual and automatic testing, we find multiple temporal errors and characterize the conditions under which QA performance deteriorates.
Our analysis contributes to understanding LLM limitations and offers valuable insights into developing future models that can better cater to the demands of temporally-oriented tasks. The code is available\footnote{https://github.com/jwallat/temporalblindspots}.   
\end{abstract}

%% file: introduction.tex
\section{Introduction}
\label{sec:intro}

Autoregressive large language models (LLMs) like~\cite{brown:2020:neurips:gpt3,Touvron:2023:arxiv:LLaMa} are versatile and a general-purpose solution to many natural language processing tasks. 
These models, benefiting from advanced natural language understanding capabilities, have demonstrated impressive zero-shot and few-shot performance. 
However, both pre-training and instruction-tuning data utilized in LLMs are often confined to a specific corpus and instruction demonstrations.
However, there is a poor understanding of how much these large language models exhibit temporal knowledge and orientation in time.
Consequently, this raises concerns regarding the effectiveness of LLMs for a range of tasks involving temporal intents like question answering and search over historical sources~\cite{DBLP:conf/sigir/WangJY22}, QA over legal and personal temporal collections~\cite{qin:2020:www:ltrpersonalsearch,gupta:2019:www:personalizedspellcorrectionPersonalSearch}, or fact checking~\cite{nakov:2021:ijcai:automaticFactChecking}.

\input{Tables/intro_example_new}

Extensive research work towards understanding the capabilities and limitations of LLMs has focussed on knowledge probing~\cite{DBLP:conf/emnlp/PetroniRRLBWM19,kassner:2021:eacl:multilinguallama}, adversarial examples~\cite{ganguli:2022:arxiv:redteamingLMs}, and risk analysis~\cite{derczynski:2023:arxiv:riskcards} among others.
Probing methods have focussed on factual~\cite{kassner:2021:eacl:multilinguallama, west:2022:acl:probingfactualgrounded}, common-sense ~\cite{talmor:2019:naacl:commonsenseQA}, social~\cite{yin:2022:emnlp:geodiversecommonsense}, numerical~\cite{lin:2020:acl:numericalcommonsenseknowledge} and spatial~\cite{cohn:2023:arxiv:spatialcommonsense} knowledge. 
However, most of the probing methods utilize bi-directional and non-instruction-tuned models.
We do not focus on adversarial examples~\cite{ganguli:2022:arxiv:redteamingLMs} but on natural questions that are incorrectly answered due to a lack of temporal understanding.
Unlike the existing work, we analyze the temporal knowledge blind spots and temporal understanding of instruction-tuned LLMs for natural questions with a temporal intent. 
A few examples of temporally-induced errors on our datasets are shown in Table~\ref{tab:intro_example}.
To the best of our knowledge, our study is the first to inspect LLMs, their temporal blind spots, and their abilities in navigating over time.

\subsection{Aim of this study}

In this study, we aim to investigate the underlying limitations of general-purpose LLMs when deployed for tasks that require temporal knowledge and temporal understanding. 
Our research focuses on evaluating the abilities of LLMs in the context of temporal knowledge retrieval and processing and involves comprehensive testing over three temporal question-answering datasets: \textit{TemporalQuestions}~\cite{wang:2021:IRJ:qatemporalcollections}, \textit{ArchivalQA}~\cite{DBLP:conf/sigir/WangJY22}, and \textit{TempLAMA}~\cite{dhingra:2022:tacl:lmsastkbs}. 

We pay particular attention to the handling of factual temporal knowledge and the processing of complex temporal information. Specifically, we experiment with different time-referencing schemes (absolute and relative), as well as experiment with the corruption of time references.  
Through this examination, we hope to shed light on the extent to which LLMs can effectively address the challenges associated with temporal knowledge management.

%% file: Tables/intro_example_new.tex
\begin{table}[ht]
\center
\small
\begin{tabular}{lll}
\toprule
\multicolumn{3}{l}{\texttt{Which film won seven Oscars in 1994?}}             \\
                 \midrule
\textsc{ChatGPT} & \textcolor{magenta}{\textsc{Forrest Gump}}     & \xmark{} \textit{Temporal shift:} \\
Correct          & \textcolor{teal}{\texttt{Schindler's List}} &F.G. won Oscars in \textbf{1995}    \\
\midrule
\\
                \multicolumn{3}{l}{\makecell[l]{\texttt{Who lost the WBA boxing title, refusing to fight Tony } \\ \texttt{Tucker in March 1995?}}}      \\
                 \midrule
\textsc{Alpaca}  & \textcolor{magenta}{\texttt{Mike Tyson}}       & \xmark{} \textit{Time invariant:}        \\
Correct          & \textcolor{teal}{\texttt{George Foreman}}   &   time disregarded \\
\midrule
\\
                  \multicolumn{3}{l}{\texttt{Tom Brady played for which team in 2020?}} \\
                 \midrule
\textsc{Alpaca}  & \textcolor{magenta}{\texttt{New England Patriots}} & \xmark{} \textit{Temporal inertia}: Brady   \\
Correct          & \textcolor{teal}{\texttt{Tampa Bay Buccaneers}} &  joined Buccaneers in 2020   \\
\bottomrule
\end{tabular}
\caption{Examples of temporal blindspots in LLMs. }
\label{tab:intro_example}
\end{table}

%% file: related-work.tex
\section{Related Work}
\label{subsec:bio}

\subsection{Temporal Aspects of Text}
Several NLP and IR tasks have benefited from utilizing either of the two key temporal dimensions of texts (be it documents or queries), which are: \emph{creation time} and \emph{focus time} (aka. content time) \cite{kanhabua:2016:acm:TIR}.
The former refers to when a text was created (e.g., document timestamp or query issuing time), while the latter is the time mentioned or implicitly referred to in text, e.g., a document about WWII has the focus time of $1939-1945$. Similarly, the focus time of a query "Winter Olympics 1988" would refer to when this sports event took place in Calgary (Feb 13 - Feb 28, 1988), and if this query were issued today, then its creation time would be the current day. 

Temporal expressions embedded in a text may be relative, implicit, or underspecified, making their proper disambiguation challenging. In practice, texts may also contain sentences or paragraphs related to different time points (e.g., referring to multiple past events from different time periods). Hence, a document focus time may need to be represented as a set of time intervals \cite{jatowt:2013:CIKM:estimatingfocustime}. 
Note that focus time is not always explicitly mentioned in the form of temporal expressions, which temporal taggers like SuTime \cite{chang:2012:lrec:sutime} or HeidelTime \cite{strotgen:2010:acl:heideltime} could extract (e.g., mentions of historical events from WWII without specifying any dates, or a query such as "Winter Olympics in Calgary"), in which case particular reasoning approaches may be required to anchor text over time dimension \cite{jatowt:2013:CIKM:estimatingfocustime}. 

Exploiting the above-mentioned two kinds of temporal information has been gaining increased importance in temporal information retrieval and NLP, and their inter-relations have been utilized to develop various time-specific methods and applications \cite{holzmann:2016:WWW:tempas,holzmann:2016:SIGIR:capabilities}. 
They have been applied to query and document matching in time-aware document ranking \cite{gupta:2014:CIKM:timeintervals,singh:2018:axiv:history} or temporal search in web archives \cite{anand:2012:SIGIR:indexmaintainance}. More recently, documents' publication and content time have been harnessed in temporal question answering over longitudinal document collections \cite{wang:2021:IRJ:qatemporalcollections} where questions could be time-scoped (contain explicit time expressions) or the time can be implicit. Other related works include search results diversification \cite{berberich2013temporal}, clustering \cite{svore:2012:SIGIR:temporallywebsnippets}, summarization \cite{barros:2019:IRJ:natsum}, document timestamping \cite{wang2021event}, and event ordering \cite{honovich:2020:ACL:MRhistoricalevents}. Additionally, Yamato et al. \cite{yamamoto:2008:WISE:supportingjudgement} explored changes in web facts' popularity over time, and Joho et al. \cite{joho:2013:WWW:temporalwebexperience} investigated temporal aspects in user search intents during web searches.

\subsection{Time \& Pre-trained Language Models}
BERT \cite{devlin:2019:naacl:bert} was one of the first breakthroughs contributing to the recent success of pre-trained language models. Its two pre-training tasks \emph{masked language modeling} and \emph{next sentence prediction} are time-agnostic, resulting in relatively poor temporal reasoning abilities of BERT and other subsequent models. Inspired by Salient Span Masking \cite{guu:2020:PMLR:RAGpretraining} and entity replacement incorporated into pretraining tasks \cite{xiong:2020:openreview:pretrainedencyclopedia}, some researchers proposed to improve the performance of language models on various domain-specific tasks (e.g., entity-related tasks) by experimenting with the adaptation of pre-training tasks to specialize models for certain types of knowledge. 
Althammer et al. \cite{althammer:2021:arixv:linguisticallymasking} proposed linguistically informed masking, demonstrating that paying attention to complex linguistic expressions benefits downstream tasks related to patent document processing. 

Others have recently explored incorporating temporal knowledge into language models \cite{dhingra:2022:tacl:lmsastkbs, rosin2022temporal, rosin2022time}. 
A simple modification to pre-training that parametrizes MLM objective with timestamp information using temporally-scoped knowledge has been proposed in \cite{dhingra:2022:tacl:lmsastkbs} with the experiments conducted on a downstream task of question answering. Specifically, the authors focus on temporally scoped facts (e.g., "Cristiano Ronaldo played for Real Madrid in 2012", "Cristiano Ronaldo played for Juventus FC in 2019"), modifying the pretraining by transforming the input form with the prefixed temporal information (e.g., "year:2012 text: Cristiano Ronaldo plays for X" and "year:2019 text: Cristiano Ronaldo plays for X"). 

Rosin and Radinsky \cite{rosin2022temporal} and Rosin et al. \cite{rosin2022time} focus on the task of semantic change detection and characterization (i.e., detection of words that underwent semantic drift and calculation of the drift's extent), achieving a relatively modest improvement. The solution of Rosin and Radinsky \cite{rosin2022temporal} is to extend the self-attention mechanism by incorporating timestamp information to compute new attention scores, while the one of Rosin et al. \cite{rosin2022time} relies on training BERT by concatenating timestamp and text as input. The authors of \cite{rosin2022time} experiment also with the sentence time prediction task. Yet, their solution performs inferior to the fine-tuned BERT model.

More recently, Cole et al. \cite{cole2023salient} incorporated content time with the transformer encoder-decoder architecture (T5 model), masking the content time and experimenting with various temporal tasks.
Lastly, Wang et al. \cite{Wang2023} exploit both the creation and focus time during pre-training with transformer encoder-only architecture on a temporal news collection, experimenting on a range of tasks including temporal information retrieval, question answering over temporal collections, document timestamping, and event dating.

\input{Tables/datasets_overview}
 
\subsection{Temporal Knowledge Datasets}

Quite a large number of question answering benchmarks have been introduced recently \cite{baradaran2020survey, dzendzik2021english}. The datasets for querying time-related factual knowledge are, however, relatively less common. NewsQA \cite{trischler-etal-2017-newsqa} is a machine reading comprehension dataset containing $119K$ text span answers created based on a collection of CNN news articles published over nine years (2007 - 2015). Questions in NewsQA require additional background knowledge from the original paragraphs, based on which they were generated, to be understood and correctly answered. Thus, they cannot be considered as forming a standalone open QA dataset.  

\textit{ArchivalQA} \cite{DBLP:conf/sigir/WangJY22} is a large-scale open domain question answering dataset containing over half a million questions created from the New York Times archive \cite{sandhaus2008new} - a news article dataset that has been frequently used for temporal information retrieval \cite{kanhabua:2016:acm:TIR, campos:2014:acm:SurveyTIR}. 
The \textit{TemporalQuestions} dataset proposed in \cite{wang:2021:IRJ:qatemporalcollections} covers the same time period. However, its questions concern relatively major and well-known events. For both ArchivalQA and TemporalQuestions, half of the questions are time-scoped, while the other half do not contain any temporal expressions. 
Dhingra et al. \cite{dhingra:2022:tacl:lmsastkbs} proposed \textit{TempLAMA} - a dataset of temporally-scoped knowledge probes based on collecting subject-object relations from the 2020 Wikidata snapshot. The dataset contains over 50k data instances whose subjects and objects are both entities with their Wikipedia pages and which are in the form of manually written template cloze queries.

\textit{TempQuestions}~\cite{TempQuestions} contains 1,271 questions obtained by selecting time-related questions from other datasets\footnote{Free917, WebQuestions and ComplexQuestions datasets} with additional curation and tagging of temporal cues.
Another dataset, TimeSensitiveQA \cite{DBLP:journals/corr/abs-2108-06314}, provides about 40k questions of temporal nature obtained from wikidata after identifying time-evolving facts and considering four identifying common reasoning types ("in," "between," "before," "after"), requiring either event ordering or event locating information.
Additionally, there is a growing number of datasets devoted to temporal commonsense reasoning \cite{emnlp2023,wenzel:2023:arxiv:Tcommonsensereasoning} rather than to asking about factual temporal knowledge; this task is, however, outside of the scope of our study. Finally, Temporal Information Retrieval datasets like \cite{joho2014ntcir} contain queries with their temporal search intents (e.g., past, future, temporal, or present).

\subsection{Examining Abilities of LLMs}

Existing literature on inspecting different types of knowledge contained in language models has been extensively researched in the last five years in the context of explainability and interpretability~\cite{anand2022explainable,DBLP:conf/emnlp/PetroniRRLBWM19}. 
Petroni et al. \cite{DBLP:conf/emnlp/PetroniRRLBWM19} first introduced the notion of \textit{knowledge probing} that checked the ability of LMs to store factual knowledge about entities. Subsequently, many probing methods have been proposed to elicit factual~\cite{kassner:2021:eacl:multilinguallama, west:2022:acl:probingfactualgrounded}
or commonsense knowledge~\cite{talmor:2019:naacl:commonsenseQA}, including social~\cite{yin:2022:emnlp:geodiversecommonsense}, numerical~\cite{lin:2020:acl:numericalcommonsenseknowledge} and spatial~\cite{cohn:2023:arxiv:spatialcommonsense} knowledge. Jain et al. \cite{emnlp2023} analyzed temporal commonsense reasoning capabilities of LLMs on a range of datasets such as whether the models can correctly find typical time for certain actions, typical duration intervals, or common order.
Different from existing work, we analyze the temporal knowledge, blind spots, and temporal understanding of LLMs. 
Other partially related lines of research suggest developing task-specific riskcards~\cite{derczynski:2023:arxiv:riskcards} for structured evaluation of LM risks in deployment scenarios. 
To the best of our knowledge, our study is the first to examine the temporal knowledge and abilities of LLMs. Existing research either takes an adversarial approach~\cite{perez:2022:emnlp:redteamingLMswithLMs}, focuses on specific risks~\cite{derczynski:2023:arxiv:riskcards}, or does not center on LLMs~\cite{DBLP:conf/emnlp/PetroniRRLBWM19, kassner:2021:eacl:multilinguallama, west:2022:acl:probingfactualgrounded}.

%% file: Tables/datasets_overview.tex
\begin{table*}[ht!]
 \small
\begin{tabular}{lllcll}
\hline
\textbf{Dataset} & \textbf{Example Question} & \textbf{Answer} & \textbf{\#Qs} & \textbf{Scope} & \textbf{Type}\\ \hline
\textbf{TemporalQuestions} & \texttt{Who did President Bush run against in 2004?}      & John Kerry           & 1,000 &1987-2007 & major events \\
\textbf{ArchivalQA }       & \texttt{What was Ankara's official aid bill for in 1997?} & Cyprus               & 60,000 &1987-2007 & detailed, news \\
\textbf{TempLAMA}          & \texttt{Cristiano Ronaldo played for which team in 2020?}   & Juventus FC & 50,310 & 2010-2020 & detailed, entities \\ \hline
\end{tabular}
\caption{Overview of the temporal datasets used in this study.}
\label{tab:datasets_overview}
\end{table*}

%% file: study.tex
\section{Study Details}
\label{sec:format}

\subsection{Models}
\subsubsection{\alpaca{}}
The \alpaca{} model\footnote{https://github.com/tatsu-lab/stanford\_alpaca} is an instruction-tuned derivative of the LLaMa foundation model~\cite{Touvron:2023:arxiv:LLaMa}. The LLaMa model has been trained on the CommonCrawl\footnote{http://www.commoncrawl.org/}, C4~\cite{Raffel:2020:JMLR:T5_C4}, Github, August 2022 Wikipedia\footnote{https://en.wikipedia.org/} dump, ArXiv, and StackExchange data. 
Then, the LLaMa model was trained to follow instructions using a set of 52k examples generated from OpenAI's \dvc{} API. We use the smaller version with 7 billion parameters. 

\subsubsection{\dvc{}}
OpenAI's \dvc{}\footnote{https://platform.openai.com/docs/models/gpt-3-5} is a LLM from the GPT-3~\cite{brown:2020:neurips:gpt3} family. It was built from the InstructGPT model \cite{Ouyang:2022:neurips:InstructGPT} and consists of 175B parameters. A mixture of training datasets has been used: a filtered version of CommonCrawl, an expanded version of WebText~\cite{Radford2019LanguageMA}, two not further specified internet-based book corpora, and the English Wikipedia. The most recent information in \dvc{}'s training data is from June 2021. To probe the model's parametric memory, we adapt the OpenAI playground's\footnote{https://platform.openai.com/playground} default QA prompt to achieve shorter answers. For completeness, we add examples of our prompts in the Appendix.

\subsubsection{Open-Source LLMs}
We further use multiple openly available instruction-tuned LLMs. \openllama{}\footnote{https://huggingface.co/VMware/open-llama-7b-v2-open-instruct} \cite{openlm2023openllama} was trained on a mixture of Falcon Refined-Web~\cite{penedo:2023:arxiv:falconrefinedweb}, Starcoder~\cite{li:2023:arxiv:starcoder} and parts of the RedPajama dataset~\cite{together2023redpajama}. \rpsmall{}\footnote{https://huggingface.co/togethercomputer/RedPajama-INCITE-Instruct-3B-v1} and \rpbig{}\footnote{https://huggingface.co/togethercomputer/RedPajama-INCITE-7B-Instruct} were trained on 1.5T token of the RedPajama dataset~\cite{together2023redpajama}, and \falcon{}\footnote{https://huggingface.co/tiiuae/falcon-7b} \cite{falcon40b} was trained on Falcon Refined-Web as well as a mixture of curated corpora. Since the same instruction format was used to train \openllama{} and \alpaca{}, we use the identical prompt for both models. The remaining models will be tested using the \dvc{} prompt, as shown in the Appendix. 

\subsection{Datasets}

This study uses three datasets containing time-scoped questions: TemporalQuestions \cite{wang:2021:IRJ:qatemporalcollections}, ArchivalQA \cite{DBLP:conf/sigir/WangJY22}, and the TempLAMA dataset \cite{dhingra:2022:tacl:lmsastkbs}. The overview of the three datasets, selected examples, and further information is given in Table~\ref{tab:datasets_overview}.

\subsubsection{TemporalQuestions.}
TemporalQuestions dataset \cite{wang:2021:IRJ:qatemporalcollections} contains 1,000 human-generated questions about major events where half is explicitly and half implicitly time-scoped, meaning that half the questions contain temporal expressions while the remaining questions lack any temporal references. The questions were carefully selected from several history quiz websites, existing QA datasets such as SQUAD 1.1 \cite{rajpurkar2016squad} and TempQuestions~\cite{TempQuestions}, or manually created from Wikipedia's year pages\footnote{E.g., \url{https://en.wikipedia.org/wiki/1989}}.

\subsubsection{ArchivalQA}
The ArchivalQA train split of time-scoped questions, which we use in this analysis, was generated from the NYT News Corpus using T5-base model \cite{Raffel:2020:JMLR:T5_C4} fine-tuned on SQUAD 1.1 \cite{rajpurkar2016squad}. The questions were subsequently subject to multiple stages of systematic filtering, including syntactic filtering (e.g., dropping too short/long questions, questions with answers embedded in their content, ones with unresolved pronouns, and so on) as well as filtering questions that are not specific enough (overly general questions) and temporally ambiguous ones (i.e., questions with multiple possible answers over time) based on the application of several dedicated classifiers that were trained on specially prepared datasets. 

\input{Tables/all_results}

ArchivalQA mostly contains detailed questions about the past, often on minor events, focussing on the period of 1987-2007, i.e., the time scope covered by the NYT corpus. Like TemporalQuestions, ArchivalQA contains a mixture of questions, including absolute time references and questions lacking any temporal expressions.

\subsubsection{TempLAMA}
The TempLAMA dataset is a set of KG triples for nine relations, such as "plays for" or "head of government." It covers more recent information from 2010 to 2020 than the other two datasets. 
While the first two datasets are distributed as question-answering datasets, TempLAMA is designed as a cloze task. Therefore, we reformulated the nine relations used in TempLAMA to actual question-answering pairs as described in the Appendix.

\subsubsection{Evaluation}
We evaluate the generated responses with the standard exact match (EM) and F1 score. Additionally, since the LLMs in our experiments tended to generate longer answers (leading to lower EM/F1 scores), we included a "contains" metric. This metric utilizes string matching to check if the answer is contained in the generated text (after removing punctuation and lowercasing). Lastly, we report the BERT-based answer equivalence metric (BEM) \cite{bulian:2022:emnlp:BEM}. BEM utilizes a BERT model fine-tuned to detect equivalent answers by semantic (and not token) matching.

%% file: Tables/all_results.tex
\begin{table*}[ht!]
\begin{tabular}{lcccc|cccc|cccc}
\hline
\textbf{Model} & \multicolumn{4}{c|}{TemporalQuestions} & \multicolumn{4}{c|}{ArchivalQA} & \multicolumn{4}{c}{TempLAMA} \\ \cline{2-13}
               & BEM     & Cont.    & F1      & EM      & BEM    & Cont.  & F1    & EM    & BEM   & Cont.  & F1   & EM   \\ \hline
\dvc{}        & \textbf{75.6}        & \textbf{66.8}     & \textbf{64.0}      & \textbf{52.0}      & \textbf{30.5}       & \textbf{21.7}   & \textbf{20.7}  & \textbf{10.7}  & \textbf{29.6}      & \textbf{22.2}   & \textbf{30.7} & \textbf{16.3} \\
\alpaca{}         & 57.1        & 46.2     & 37.2    & 26.6    & 30.0       & 16.5   & 11.0  & 4.4   & 28.2      & 15.1   & 16.4 & 3.8  \\
\openllama{}      & 28.7    & 23.1     & 22.9    & 16.5    & 14.7   & 9.4    & 7.8   & 3.3   & 12.0  & 7.0    & 12.7 & 3.5  \\
\falcon{}       & 32.6    & 26.4     & 26.2    & 16.2    & 13.4      & 7.5       & 8.7       & 2.8      & 10.5      & 6.5       & 12.8    & 3.0     \\
\rpbig{}   & 42.6    & 35.1     & 40.2    & 33.7    & 14.5       & 9.3       & 11.1      & 6.4      & 15.7      & 10.6        & 19.0      & 11.0     \\
\rpsmall{}   & 27.1    & 22.5     & 25.8    & 20.3    & 12.4       & 7.2       & 8.4      & 3.9      & 11.4      & 7.0       & 10.8     & 4.6     \\ \hline
\end{tabular}
\caption{Overall results of our models on the three temporal datasets - TemporalQuestions, the time-split of ArchivalQA, and the reformulated TempLAMA examples.}
\label{tab:results_all}
\end{table*}

%% file: results.tex
\section{Analysis}

\subsection{Temporal Knowledge of LLMs}

First, we assess the general performance of the models on our temporal datasets.
The results of several LLMs on the three datasets in this study can be seen in Table~\ref{tab:results_all}.
Not surprisingly, we find the much bigger \dvc{} to perform relatively better than the smaller models. 
We, however, generally observe relatively poor performance of our models, even on rather simple questions about major events from the past (TemporalQuestions).
On the two other datasets, the performance of the LLMs is generally more lacking, and the numbers are quite low. As mentioned, ArchivalQA contains more fine-grained questions, i.e., ones about relatively minor or detailed events, than TemporalQuestions, while both cover the same time period. In the context of the smaller open-source models, we observe \rpbig{} to perform relatively well among those models at answering temporally scoped questions.

\paragraph{\textbf{Insight.}} 
The results suggest that LLMs exhibit limited ability in answering questions about the past and, overall, seem to lack knowledge regarding specific details of past events.

\subsection{Do LLMs Prioritize Recent Knowledge?}

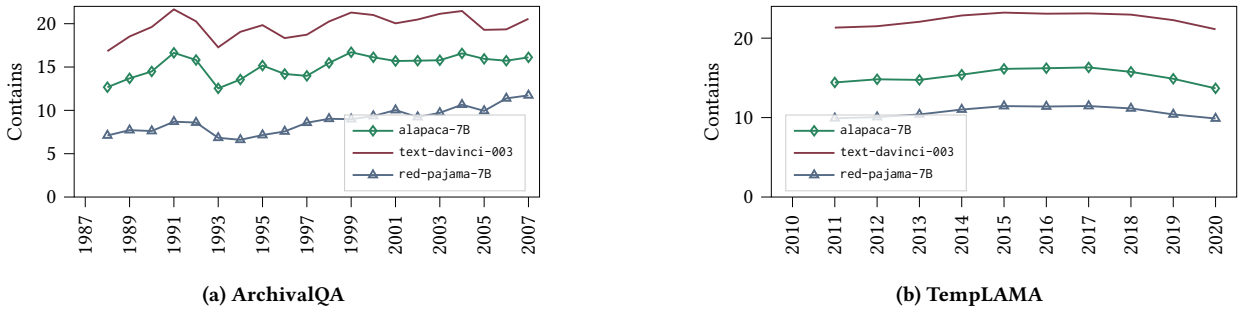
\begin{figure*}
     \begin{subfigure}[b]{0.48\textwidth}
        \begin{center}
            \scalebox{.9}{\input{Figures/time_strat/archivalqa_line_contains}}
        \end{center}
        \caption{ArchivalQA}
        \label{fig:time_strat_archivalqa}
     \end{subfigure}
     \hfill
     \begin{subfigure}[b]{0.48\textwidth}
        \begin{center}
            \scalebox{.9}{\input{Figures/time_strat/templama_line_contains}}
        \end{center}
        \caption{TempLAMA}
        \label{fig:time_strat_templama}
     \end{subfigure}
     \caption{Stratification of the \alpaca{}, \rpbig{}, and \dvc{} models on the ArchivalQA and TempLAMA datasets. Stratified by years, the trendline is the moving average with a window of 2. We do not show plots for the TemporalQuestions dataset since the dataset is not large enough for computing individual results per year.}
     \label{fig:time_strat}
\end{figure*}

To further understand the performance of the LLMs on temporal datasets, we investigate whether the LLMs prioritize temporally newer over older knowledge. In other words, we wish to analyze the effect of the passage of time on the quality of answers. Other than existing related work (e.g., \cite{lazaridou:2021:NeurIPS:mindthegap}), we do not investigate the model's behavior on new text produced after the model was trained, but how the model remembers historical information. To do so, we select those questions from the $60k$ questions of ArchivalQA that contain absolute year references. 
Next, we stratify the performance based on the years of the resulting set of around $27k$ QA pairs (Figure~\ref{fig:time_strat}\footnote{Similar trends across other models and metrics are reported in the Appendix.}).  
We see a slight trend of improvement over time, with the best results for the more recent years (2005-2007).
This is somewhat expected and could be likely due to the forgetting effect (declining remembering of older years compared to the remembering of recent years), the trend that was also observed in large news article collections \cite{DBLP:conf/cikm/YeungJ11} and collections of tweets related to history \cite{DBLP:journals/jodl/SumikawaJ21}.

Given that ArchivalQA does not cover the latest years, we next investigate the results from $2010$ to $2020$ repeating the same experiment on TempLAMA, keeping in mind that our models were trained on information up until late 2021 (\dvc{}), mid $2022$ (\alpaca{}), 2023 (others). The stratified per-year performance on the TempLAMA dataset can be seen in Figure~\ref{fig:time_strat_templama}. 
Here, we also observe a trend of more recent information being better captured until the performance peaks in 2015 and 2017, respectively. Then, a decrease is seen for more recent information (2017-2020). We hypothesize that this recent decline could be due to older facts being more prevalent in the training data. At the same time, the newer information may not yet be sufficiently common in those data to be preferred by the models when answering questions on the latest events. Such \textit{temporal inertia} can be expected as information within large document collections like the web is not being changed instantaneously at all relevant places but is rather gradually becoming updated.
Furthermore, we believe that the models do not correctly model the temporality of the training data.
If the models correctly recognize and adequately utilize the temporal signals (both publication and content time) in the documents used for training, they could prefer more recent information over older ones. In such a somewhat ideal case, even a few documents (or, in theory, even a single reliable document) should be enough for the models to update their parametric knowledge with more recent facts and thus correctly answer questions on dynamic topics.

\paragraph{\textbf{Insight.}} 
LLMs seem to capture more recent information better than older information. 
However, such a pattern appears to occur only up to a certain point, as we observe in the case of TempLama (2017). Temporal inertia could be a possible explanation here, such that the most recent information is still not sufficiently prevalent in the training data of language models, while the models do not correctly distinguish between up-to-date and obsolete knowledge.

\subsection{Relative vs. Absolute Time References}
Next, we investigate to what extent the models are sensitive to the type of time information contained in questions. We start by comparing the usefulness of relative and absolute time references. We first experiment with absolute temporal expressions, but we later convert them into relative ones to see if there would be any change in results. 
As in the latter case, the temporal distances are now made to be relative, 
we expect the models to perform worse. This is because resolving back relative expressions to their absolute form requires certain rudimentary calculations, which we do not expect the models to always perform correctly (or to perform at all). On the other hand, if reasoning and knowledge retrieval done by the models were based purely on relative temporal expressions (without converting them to the absolute form), the models would not be effective either since the actual meaning of a relative temporal expression depends on the publication date of its containing document\footnote{Or, in some cases, on absolute time expressions stated earlier within the content of the document.} - the type of information that does not seem explicitly utilized in the current mainstream LLMs.

\begin{figure}
     \begin{subfigure}[b]{0.236\textwidth}
        \begin{center}
            \scalebox{.89}{\input{Figures/referencing/relative_absolute_time_alpaca}}
        \end{center}
        \caption{\alpaca{}}
     \end{subfigure}
     \begin{subfigure}[b]{0.236\textwidth}
        \begin{center}
            \scalebox{.89}{\input{Figures/referencing/relative_absolute_time_davinci}}
        \end{center}
        \caption{\dvc{}}
     \end{subfigure}
     \caption{Relative and absolute time referencing.}
     \label{fig:relative_absolute_time}
\end{figure}
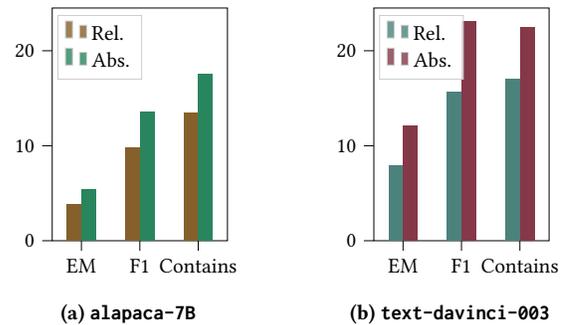

Given a question such as "Who was the American president in 2018?" we then change the absolute reference of "2018" to a relative reference ("3 years ago"). Since both models responded 2021 when asked for which year it is now\footnote{Asking with text: "What year do we have?" using our prompts (Appendix).}, we use 2021 as a reference year to compute all the relative references. The results on the subset of 27k ArchivalQA QA pairs that originally contained absolute year references can be seen in Figure~\ref{fig:relative_absolute_time}.

We observe that the performance drops by 23-30\% and 24-35\% for \alpaca{} and \dvc{}, respectively.
This might be due to relative references requiring an additional reasoning step. However, through qualitative tests, we have found the models (\dvc{}, \alpaca{}) to be able to perform this reasoning when asked explicitly\footnote{We asked: "If today is 2021, what year was 4 years ago?"}. Yet, it seems that the models prefer absolute time references in arbitrary questions - potentially to match the same absolute reference tokens in the training data. 

Given the studied language models have been trained without explicitly considering the publication time information, relative references in the training data are then less trustworthy (a mention of "2 years ago" is only useful when one knows when the document containing it was written), the models might learn to disregard these references and focus on matching absolute ones. Therefore, we find (relative) \textit{time referencing} to be a blind spot in LLMs.  

\paragraph{\textbf{Insight.}} LLMs perform up to 35\% worse when given relative time references (questions referring to time such as "4 years ago") than when absolute date expressions are used. 
Users are advised to use absolute temporal references when querying current LLMs.

\subsection{How Important is Time Referencing?}
Given that the performance of the models varies with the type of time expressions, we further investigate the importance of time references. 
To do so, we corrupt the time references by replacing a reference with a random year or by deliberately changing it to be off by a specific number of years. 

\subsubsection{Does Corruption of Time References has any Effect?}
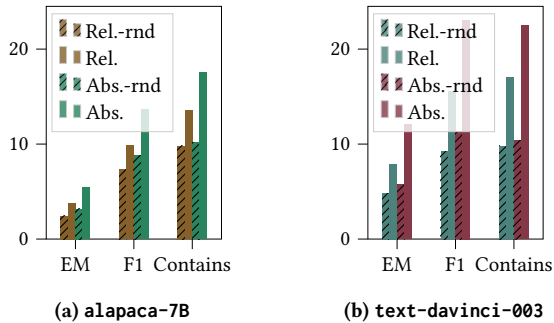
\begin{figure}
     \begin{subfigure}[b]{0.236\textwidth}
        \begin{center}
            \scalebox{.89}             {\input{Figures/referencing/absolute_relative_random_reference_alpaca}}
        \end{center}
        \caption{\alpaca{}}
     \end{subfigure}
     \begin{subfigure}[b]{0.236\textwidth}
        \begin{center}
            \scalebox{.89}{\input{Figures/referencing/absolute_relative_random_reference_davinci}}
        \end{center}
        \caption{\dvc{}}
     \end{subfigure}
     \caption{Effect of randomized relative and absolute time references. Textured bars show the randomized variants.}
     \label{fig:absolute_relative_random_reference}
\end{figure}

First, Figure~\ref{fig:absolute_relative_random_reference} shows the performance of \dvc{} and 
the \alpaca{} models given absolute and relative time references with both the correct and random years. For these experiments, we sample 3k QA pairs from the 27k ArchivalQA questions containing an absolute year reference. For the random setting, we replace the years with random years between 1900 and 2021. 

After randomizing the time references, we observe the model performance to drop by an average of 26\% and 44\% (relative) as well as 40\% and 53\% (absolute) for \alpaca{} and \dvc{}, respectively. Interestingly, randomizing the absolute time references has a similar impact on the model performance as when switching from absolute to relative references. However, randomizing the relative time references still results in decreased performance, indicating that relative references are not entirely useless or an active distraction.

\paragraph{\textbf{Insight.}} Randomizing the time reference leads to a performance decrease of 40\% (\alpaca{}) and 53\% (\dvc{}).

\input{Figures/manual-analysis}
\subsubsection{How Robust are LLMs Under Wrong Time References?}Lastly, we more closely analyze the effect of errors in temporal references. In real life, users may misremember or confuse years; hence, such cases are not completely unrealistic. To do so, we corrupt the time references of the 27k ArchivalQA questions containing years to be wrong by $X$-years and observe the impact on model performance. That is, we change the time references by $X$ years (e.g., 2014 to 2011). The results are shown in Figure~\ref{fig:absolute_relative_time_all}.

\input{Figures/corrupted-time}

Slight imprecisions in the time references only have minor effects. When asking the model about the past, being off by three years reduces the model's performance by 3\% (relative reference) and 10\% (absolute reference).
More significant deviations lead to performance converging towards the performance seen with randomized references. Corrupting the time expression to be off by 20 years reduces \alpaca{}'s performance by 33-35\% for relative and absolute expressions. We note that the performance degrades with the degree of the introduced error, suggesting that it is harder for the models to recover from larger errors. When comparing the effect of corrupting the references to removing the entire time reference, we observe that relative references distract the model. 

\paragraph{\textbf{Insight.}} 
The amount of error in the time references embedded in questions appears to correlate with the answer quality. The larger the introduced error, the worse the results. Not using time references at all seems to be preferential over relative references.

%% file: Figures/time_strat/archivalqa_line_contains.tex
\begin{tikzpicture}

    \definecolor{darkgray176}{RGB}{176,176,176}
    \definecolor{steelblue31119180}{RGB}{31,119,180}
    \definecolor{lightgray204}{RGB}{204,204,204}
    
    \begin{axis}[
    width=.99\textwidth,
    height=.2\textheight,
    legend cell align={left},
    legend style={
      fill opacity=0.8,
      draw opacity=1,
      text opacity=1,
      at={(0.03,0.97)},
      anchor=north west,
      draw=lightgray204
    },
    legend pos= south east,
    tick align=outside,
    tick pos=left,
    x grid style={darkgray176},
    xlabel={},
    xmin=-0.5, xmax=20.5,
    xtick style={color=black},
    xtick={
        0,
        2,
        4,
        6,
        8,
        10,
        12,
        14,
        16,
        18,
        20
    },
    xticklabel style={rotate=90.0},
    xticklabels={
      {1987},
      {1989},
      {1991},
      {1993},
      {1995},
      {1997},
      {1999},
      {2001},
      {2003},
      {2005},
      {2007}
    },
    y grid style={darkgray176},
    ylabel={Contains},
    ymin=0, ymax=22,
    ytick style={color=black}
    ]
    
\addplot [thick, absolute_alpaca, mark=diamond] 
table {
0           NaN
1     12.675268
2     13.693078
3     14.489479
4     16.640410
5     15.811847
6     12.538875
7     13.559208
8     15.171745
9     14.198113
10    13.982767
11    15.472181
12    16.705581
13    16.127321
14    15.695037
15    15.732949
16    15.782530
17    16.571320
18    15.943139
19    15.727453
20    16.117647
};
\addlegendentry{{\scriptsize \alpaca{}}}
\addplot [thick, absolute_davinci] 
table {
0           NaN
1     16.833546
2     18.523618
3     19.619948
4     21.652001
5     20.271034
6     17.275689
7     19.063229
8     19.831848
9     18.340223
10    18.731945
11    20.254168
12    21.287207
13    21.001716
14    20.043568
15    20.480805
16    21.139336
17    21.464653
18    19.289689
19    19.344890
20    20.564706
};
\addlegendentry{{\scriptsize \dvc{}}}
\addplot [thick, absolute_redpajama_7, mark=triangle]
table {
0           NaN
1      7.101540
2      7.716433
3      7.615371
4      8.690837
5      8.595780
6      6.835531
7      6.600993
8      7.150892
9      7.566915
10     8.572671
11     9.025224
12     8.984692
13     9.346622
14    10.032676
15     9.220571
16     9.716375
17    10.655283
18     9.938904
19    11.381301
20    11.729412
};
\addlegendentry{{\scriptsize \rpbig{}}}
    
\end{axis}

\end{tikzpicture}

%% file: Figures/time_strat/templama_line_contains.tex
\begin{tikzpicture}

\definecolor{darkgray176}{RGB}{176,176,176}
\definecolor{steelblue31119180}{RGB}{31,119,180}
\definecolor{lightgray204}{RGB}{204,204,204}

\begin{axis}[
width=.99\textwidth,
height=.2\textheight,
legend cell align={left},
legend style={
  fill opacity=0.8,
  draw opacity=1,
  text opacity=1,
  at={(0.03,0.97)},
  anchor=north west,
  draw=lightgray204
},
legend pos= south west,
tick align=outside,
tick pos=left,
x grid style={darkgray176},
xlabel={},
xmin=-0.5, xmax=10.5,
xtick style={color=black},
xtick={0,1,2,3,4,5,6,7,8,9,10},
xticklabel style={rotate=90.0},
xticklabels={
  {2010},
  {2011},
  {2012},
  {2013},
  {2014},
  {2015},
  {2016},
  {2017},
  {2018},
  {2019},
  {2020}
},
y grid style={darkgray176},
ylabel={Contains},
ymin=0, ymax=24,
ytick style={color=black}
]

\addplot [thick, absolute_alpaca, mark=diamond] 
table {
0           NaN
1     14.421616
2     14.820894
3     14.742419
4     15.399339
5     16.130400
6     16.213645
7     16.313712
8     15.756282
9     14.886085
10    13.678437
};
\addlegendentry{{\scriptsize \alpaca{}}}
\addplot [thick, absolute_davinci] 
table {
0           NaN
1     21.324557
2     21.501998
3     22.061428
4     22.852903
5     23.213109
6     23.076011
7     23.119264
8     22.958403
9     22.266788
10    21.116565
};
\addlegendentry{{\scriptsize \dvc{}}}
\addplot [thick, absolute_redpajama_7, mark=triangle] 
table {
0           NaN
1      9.911566
2     10.057162
3     10.393823
4     11.012363
5     11.444602
6     11.392969
7     11.453174
8     11.151070
9     10.399207
10     9.883215
};
\addlegendentry{{\scriptsize \rpbig{}}}
\end{axis}

\end{tikzpicture}

%% file: Figures/referencing/relative_absolute_time_alpaca.tex
\begin{tikzpicture}

\definecolor{darkgray176}{RGB}{176,176,176}
\definecolor{lightgray204}{RGB}{204,204,204}

\begin{axis}[
width=\textwidth,
height=.23\textheight,
legend cell align={left},
legend style={
  fill opacity=0.8,
  draw opacity=1,
  text opacity=1,
  at={(0.03,0.97)},
  anchor=north west,
  draw=lightgray204
},
tick align=outside,
tick pos=left,
x grid style={darkgray176},
xmin=-0.5, xmax=2.5,
xtick style={color=black},
xtick={0,1,2},
xticklabels={EM,F1,Contains},
y grid style={darkgray176},
ymin=0, ymax=24.5,
ytick style={color=black}
]
\draw[draw=none,fill=relative] (axis cs:-0.25,0) rectangle (axis cs:0,3.8);
\addlegendimage{ybar,ybar legend,draw=none,fill=relative}
\addlegendentry{Rel.}

\draw[draw=none,fill=relative] (axis cs:0.75,0) rectangle (axis cs:1,9.8);
\draw[draw=none,fill=relative] (axis cs:1.75,0) rectangle (axis cs:2,13.5);
\draw[draw=none,fill=absolute] (axis cs:0,0) rectangle (axis cs:0.25,5.4);
\addlegendimage{ybar,ybar legend,draw=none,fill=absolute}
\addlegendentry{Abs.}

\draw[draw=none,fill=absolute] (axis cs:1,0) rectangle (axis cs:1.25,13.6);
\draw[draw=none,fill=absolute] (axis cs:2,0) rectangle (axis cs:2.25,17.5);
\end{axis}

\end{tikzpicture}

%% file: Figures/referencing/relative_absolute_time_davinci.tex
\begin{tikzpicture}

\definecolor{darkgray176}{RGB}{176,176,176}
\definecolor{lightgray204}{RGB}{204,204,204}

\begin{axis}[
width=\textwidth,
height=.23\textheight,
legend cell align={left},
legend style={
  fill opacity=0.8,
  draw opacity=1,
  text opacity=1,
  at={(0.03,0.97)},
  anchor=north west,
  draw=lightgray204
},
tick align=outside,
tick pos=left,
x grid style={darkgray176},
xmin=-0.5, xmax=2.5,
xtick style={color=black},
xtick={0,1,2},
xticklabels={EM,F1,Contains},
y grid style={darkgray176},
ymin=0, ymax=24.5,
ytick style={color=black}
]
\draw[draw=none,fill=relative_davinci] (axis cs:-0.25,0) rectangle (axis cs:0,7.9);
\addlegendimage{ybar,ybar legend,draw=none,fill=relative_davinci}
\addlegendentry{Rel.}

\draw[draw=none,fill=relative_davinci] (axis cs:0.75,0) rectangle (axis cs:1,15.6);
\draw[draw=none,fill=relative_davinci] (axis cs:1.75,0) rectangle (axis cs:2,17.0);
\draw[draw=none,fill=absolute_davinci] (axis cs:0,0) rectangle (axis cs:0.25,12.1);
\addlegendimage{ybar,ybar legend,draw=none,fill=absolute_davinci}
\addlegendentry{Abs.}

\draw[draw=none,fill=absolute_davinci] (axis cs:1,0) rectangle (axis cs:1.25,23.1);
\draw[draw=none,fill=absolute_davinci] (axis cs:2,0) rectangle (axis cs:2.25,22.5);
\end{axis}

\end{tikzpicture}

%% file: Figures/referencing/absolute_relative_random_reference_alpaca.tex
\begin{tikzpicture}

\definecolor{darkgray176}{RGB}{176,176,176}
\definecolor{lightgray204}{RGB}{204,204,204}

\begin{axis}[
width=\textwidth,
height=.23\textheight,
legend cell align={left},
legend style={
  fill opacity=0.8,
  draw opacity=1,
  text opacity=1,
  at={(0.03,0.97)},
  anchor=north west,
  draw=lightgray204
},
tick align=outside,
tick pos=left,
x grid style={darkgray176},
xmin=-0.5, xmax=2.5,
xtick style={color=black},
xtick={0,1,2},
xticklabels={EM,F1,Contains},
y grid style={darkgray176},
ymin=0, ymax=24.5,
ytick style={color=black}
]
\draw[draw=none,fill=relative_rand,postaction={pattern=north east lines}] (axis cs:-0.25,0) rectangle (axis cs:-0.125,2.43);
\addlegendimage{ybar,ybar legend,draw=none,fill=relative_rand,postaction={pattern=north east lines}}
\addlegendentry{Rel.-rnd}

\draw[draw=none,fill=relative_rand,postaction={pattern=north east lines}] (axis cs:0.75,0) rectangle (axis cs:0.875,7.32);
\draw[draw=none,fill=relative_rand,postaction={pattern=north east lines}] (axis cs:1.75,0) rectangle (axis cs:1.875,9.8);
\draw[draw=none,fill=relative] (axis cs:-0.125,0) rectangle (axis cs:0,3.8);
\addlegendimage{ybar,ybar legend,draw=none,fill=relative}
\addlegendentry{Rel.}

\draw[draw=none,fill=relative] (axis cs:0.875,0) rectangle (axis cs:1,9.83);
\draw[draw=none,fill=relative] (axis cs:1.875,0) rectangle (axis cs:2,13.5);
\draw[draw=none,fill=absolute_rand,postaction={pattern=north east lines}] (axis cs:0,0) rectangle (axis cs:0.125,3.17);
\addlegendimage{ybar,ybar legend,draw=none,fill=absolute_rand,postaction={pattern=north east lines}}
\addlegendentry{Abs.-rnd}

\draw[draw=none,fill=absolute_rand,postaction={pattern=north east lines}] (axis cs:1,0) rectangle (axis cs:1.125,8.76);
\draw[draw=none,fill=absolute_rand,postaction={pattern=north east lines}] (axis cs:2,0) rectangle (axis cs:2.125,10.17);
\draw[draw=none,fill=absolute] (axis cs:0.125,0) rectangle (axis cs:0.25,5.4);
\addlegendimage{ybar,ybar legend,draw=none,fill=absolute}
\addlegendentry{Abs.}

\draw[draw=none,fill=absolute] (axis cs:1.125,0) rectangle (axis cs:1.25,13.64);
\draw[draw=none,fill=absolute] (axis cs:2.125,0) rectangle (axis cs:2.25,17.5);
\end{axis}

\end{tikzpicture}

%% file: Figures/referencing/absolute_relative_random_reference_davinci.tex
\begin{tikzpicture}

\definecolor{darkgray176}{RGB}{176,176,176}
\definecolor{lightgray204}{RGB}{204,204,204}

\begin{axis}[
width=\textwidth,
height=.23\textheight,
legend cell align={left},
legend style={
  fill opacity=0.8,
  draw opacity=1,
  text opacity=1,
  at={(0.03,0.97)},
  anchor=north west,
  draw=lightgray204
},
tick align=outside,
tick pos=left,
x grid style={darkgray176},
xmin=-0.5, xmax=2.5,
xtick style={color=black},
xtick={0,1,2},
xticklabels={EM,F1,Contains},
y grid style={darkgray176},
ymin=0, ymax=24.5,
ytick style={color=black}
]
\draw[draw=none,fill=relative_davinci_random,postaction={pattern=north east lines}] (axis cs:-0.25,0) rectangle (axis cs:-0.125,4.77);
\addlegendimage{ybar,ybar legend,draw=none,fill=relative_davinci_random,postaction={pattern=north east lines}}
\addlegendentry{Rel.-rnd}

\draw[draw=none,fill=relative_davinci_random,postaction={pattern=north east lines}] (axis cs:0.75,0) rectangle (axis cs:0.875,9.18);
\draw[draw=none,fill=relative_davinci_random,postaction={pattern=north east lines}] (axis cs:1.75,0) rectangle (axis cs:1.875,9.8);
\draw[draw=none,fill=relative_davinci] (axis cs:-0.125,0) rectangle (axis cs:0,7.9);
\addlegendimage{ybar,ybar legend,draw=none,fill=relative_davinci}
\addlegendentry{Rel.}

\draw[draw=none,fill=relative_davinci] (axis cs:0.875,0) rectangle (axis cs:1,15.56);
\draw[draw=none,fill=relative_davinci] (axis cs:1.875,0) rectangle (axis cs:2,17.00);
\draw[draw=none,fill=absolute_davinci_random,postaction={pattern=north east lines}] (axis cs:0,0) rectangle (axis cs:0.125,5.7);
\addlegendimage{ybar,ybar legend,draw=none,fill=absolute_davinci_random,postaction={pattern=north east lines}}
\addlegendentry{Abs.-rnd}

\draw[draw=none,fill=absolute_davinci_random,postaction={pattern=north east lines}] (axis cs:1,0) rectangle (axis cs:1.125,11.31);
\draw[draw=none,fill=absolute_davinci_random,postaction={pattern=north east lines}] (axis cs:2,0) rectangle (axis cs:2.125,10.4);
\draw[draw=none,fill=absolute_davinci] (axis cs:0.125,0) rectangle (axis cs:0.25,12.07);
\addlegendimage{ybar,ybar legend,draw=none,fill=absolute_davinci}
\addlegendentry{Abs.}

\draw[draw=none,fill=absolute_davinci] (axis cs:1.125,0) rectangle (axis cs:1.25,23.07);
\draw[draw=none,fill=absolute_davinci] (axis cs:2.125,0) rectangle (axis cs:2.25,22.5);
\end{axis}

\end{tikzpicture}

%% file: Figures/manual-analysis.tex
\begin{figure*}[ht]
     \begin{subfigure}[b]{0.48\textwidth}
        \begin{center}
            \scalebox{0.8}             {\input{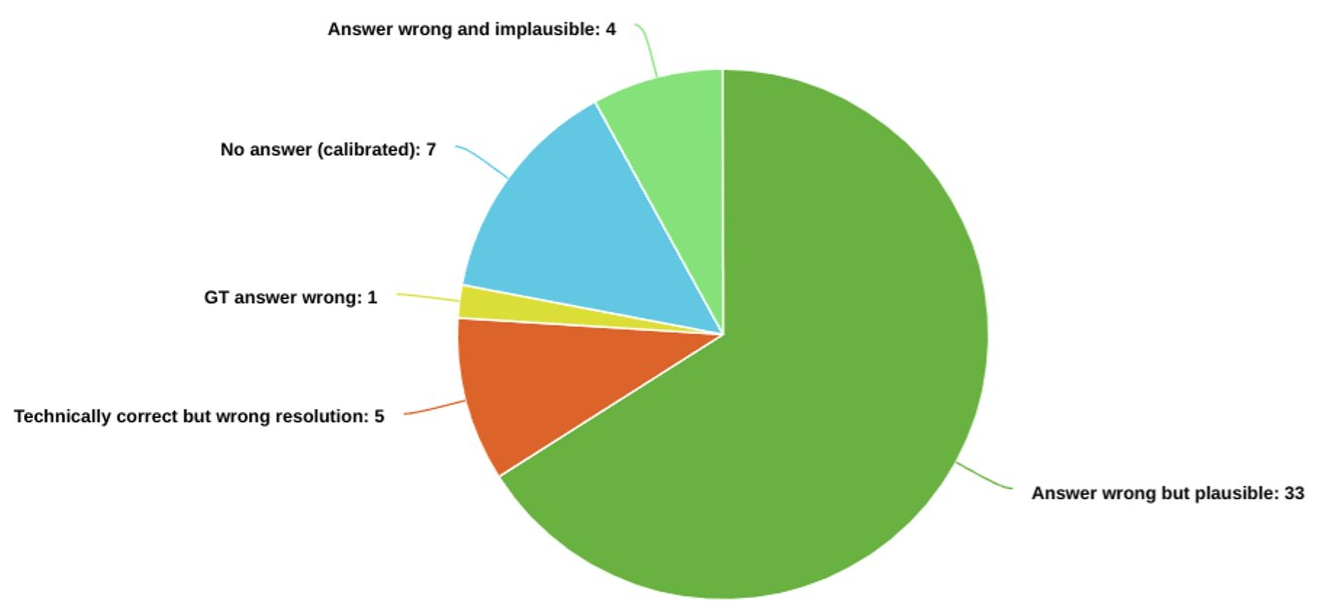}}
        \end{center}
        \caption{\alpaca{}}
     \end{subfigure}
     \hfill
     \begin{subfigure}[b]{0.48\textwidth}
        \begin{center}
            \scalebox{0.8}{\input{Figures/error_analysis/manual_analysis_davinci_archivalqa}}
        \end{center}
        \caption{\dvc{}}
     \end{subfigure}
     \caption{Manual analysis of 100 of \alpaca{}'s and \dvc{}'s wrong answers from the ArchivalQA dataset.}
     \label{fig:manual_analysis_archivalqa}
\end{figure*}

%% file: Figures/error_analysis/manual_analysis_alpaca_archivalqa.tex
\begin{tikzpicture}
    \pie[
    text={pin}, 
    /tikz/nodes={text=white, font=\bfseries},
    /tikz/every pin/.style={align=center, text=black, font=\normalfont},
    radius=1.8, 
    color={
        red!50,
        blue!60,
        magenta!70,
        cyan!80,
        teal!70,
        brown!70,
        blue!80,
        gray!90
    }
    ]{
    45/Plausible,
    4/Temporal,
    3/GT wrong, 
    23/Implausible, 
    17/No answer, 
    4/Ambiguous,
    5/Correct (other), 
    3/Granularity 
    } 
\end{tikzpicture}

%% file: Figures/error_analysis/manual_analysis_davinci_archivalqa.tex
\begin{tikzpicture}
    \pie[
    text={pin}, 
    /tikz/nodes={text=white, font=\bfseries},
    /tikz/every pin/.style={align=center, text=black, font=\normalfont},
    radius=1.8, 
    color={
        red!50,
        blue!60,
        magenta!70,
        cyan!80,
        teal!70,
        brown!70,
        blue!80,
        gray!90
    }
    ]{
        27/Plausible, 
        6/Temporal,
        6/GT wrong, 
        6/Implausible,
        46/No answer,
        3/Ambiguous,
        5/Correct (other),
        1/Granularity
    } 
\end{tikzpicture}

%% file: Figures/corrupted-time.tex
\begin{figure}[ht!]
    \begin{center}
        \scalebox{0.89}{\input{Figures/referencing/relative_time_corruption_contains}}
    \end{center}
    \caption{The effect of corrupted time of the \alpaca{} model. The relative time reference is calculated via a reference year (2021). Off-by-X means that the year is X years apart from the correct year. No time refers to the effect of entirely removing the time reference from the questions.}
    \label{fig:absolute_relative_time_all}
\end{figure}
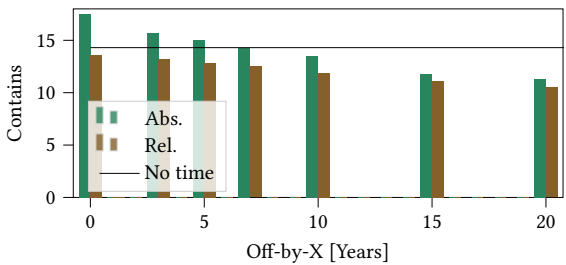

%% file: Figures/referencing/relative_time_corruption_contains.tex
\begin{tikzpicture}

\definecolor{darkgray176}{RGB}{176,176,176}
\definecolor{lightgray204}{RGB}{204,204,204}

\begin{axis}[
width=0.5\textwidth,
height=.2\textheight,
legend cell align={left},
legend style={
  fill opacity=0.8,
  draw opacity=1,
  text opacity=1,
  at={(0.03,0.97)},
  anchor=north west,
  draw=lightgray204
},
legend pos= south west,
tick align=outside,
tick pos=left,
x grid style={darkgray176},
xmin=-0.75, xmax=20.75,
xtick style={color=black},
xlabel={Off-by-X [Years]},
y grid style={darkgray176},
ylabel={Contains},
ymin=0, ymax=18,
ytick style={color=black}
]
\draw[draw=none,fill=absolute] (axis cs:-0.5,0) rectangle (axis cs:0,17.5);
\addlegendimage{ybar,ybar legend,draw=none,fill=absolute}
\addlegendentry{Abs.}

\draw[draw=none,fill=absolute] (axis cs:0.75,0) rectangle (axis cs:1,0);
\draw[draw=none,fill=absolute] (axis cs:1.75,0) rectangle (axis cs:2,0);
\draw[draw=none,fill=absolute] (axis cs:2.5,0) rectangle (axis cs:3,15.7);
\draw[draw=none,fill=absolute] (axis cs:3.75,0) rectangle (axis cs:4,0);
\draw[draw=none,fill=absolute] (axis cs:4.5,0) rectangle (axis cs:5,15.0);
\draw[draw=none,fill=absolute] (axis cs:5.75,0) rectangle (axis cs:6,0);
\draw[draw=none,fill=absolute] (axis cs:6.5,0) rectangle (axis cs:7,14.3);
\draw[draw=none,fill=absolute] (axis cs:7.75,0) rectangle (axis cs:8,0);
\draw[draw=none,fill=absolute] (axis cs:8.75,0) rectangle (axis cs:9,0);
\draw[draw=none,fill=absolute] (axis cs:9.5,0) rectangle (axis cs:10,13.5);
\draw[draw=none,fill=absolute] (axis cs:10.75,0) rectangle (axis cs:11,0);
\draw[draw=none,fill=absolute] (axis cs:11.75,0) rectangle (axis cs:12,0);
\draw[draw=none,fill=absolute] (axis cs:12.75,0) rectangle (axis cs:13,0);
\draw[draw=none,fill=absolute] (axis cs:13.75,0) rectangle (axis cs:14,0);
\draw[draw=none,fill=absolute] (axis cs:14.5,0) rectangle (axis cs:15,11.8);
\draw[draw=none,fill=absolute] (axis cs:15.75,0) rectangle (axis cs:16,0);
\draw[draw=none,fill=absolute] (axis cs:16.75,0) rectangle (axis cs:17,0);
\draw[draw=none,fill=absolute] (axis cs:17.75,0) rectangle (axis cs:18,0);
\draw[draw=none,fill=absolute] (axis cs:18.75,0) rectangle (axis cs:19,0);
\draw[draw=none,fill=absolute] (axis cs:19.5,0) rectangle (axis cs:20,11.3);

\draw[draw=none,fill=relative] (axis cs:0,0) rectangle (axis cs:0.5,13.6);
\addlegendimage{ybar,ybar legend,draw=none,fill=relative}
\addlegendentry{Rel.}

\draw[draw=none,fill=relative] (axis cs:1,0) rectangle (axis cs:1.5,0);
\draw[draw=none,fill=relative] (axis cs:2,0) rectangle (axis cs:2.25,0);
\draw[draw=none,fill=relative] (axis cs:3,0) rectangle (axis cs:3.5,13.2);
\draw[draw=none,fill=relative] (axis cs:4,0) rectangle (axis cs:4.25,0);
\draw[draw=none,fill=relative] (axis cs:5,0) rectangle (axis cs:5.5,12.8);
\draw[draw=none,fill=relative] (axis cs:6,0) rectangle (axis cs:6.25,0);
\draw[draw=none,fill=relative] (axis cs:7,0) rectangle (axis cs:7.5,12.5);
\draw[draw=none,fill=relative] (axis cs:8,0) rectangle (axis cs:8.25,0);
\draw[draw=none,fill=relative] (axis cs:9,0) rectangle (axis cs:9.25,0);
\draw[draw=none,fill=relative] (axis cs:10,0) rectangle (axis cs:10.5,11.9);
\draw[draw=none,fill=relative] (axis cs:11,0) rectangle (axis cs:11.25,0);
\draw[draw=none,fill=relative] (axis cs:12,0) rectangle (axis cs:12.25,0);
\draw[draw=none,fill=relative] (axis cs:13,0) rectangle (axis cs:13.25,0);
\draw[draw=none,fill=relative] (axis cs:14,0) rectangle (axis cs:14.25,0);
\draw[draw=none,fill=relative] (axis cs:15,0) rectangle (axis cs:15.5,11.1);
\draw[draw=none,fill=relative] (axis cs:16,0) rectangle (axis cs:16.25,0);
\draw[draw=none,fill=relative] (axis cs:17,0) rectangle (axis cs:17.25,0);
\draw[draw=none,fill=relative] (axis cs:18,0) rectangle (axis cs:18.25,0);
\draw[draw=none,fill=relative] (axis cs:19,0) rectangle (axis cs:19.25,0);
\draw[draw=none,fill=relative] (axis cs:20,0) rectangle (axis cs:20.5,10.5);

\addplot[mark=none, black, samples=2, domain=0:21] {14.3};
\addlegendentry{No time}

\end{axis}
\end{tikzpicture}

%% file: temp-errors.tex
\section{Temporal Errors}
\subsection{Error Analysis}
Given the rather low performance of our models on the temporal datasets, we now investigate the possible reasons for failures. In the first open coding step \cite{strauss:1998:qualitativeresearch}, we use 100 wrong \alpaca{} predictions to identify failure cases. Next, we manually annotate 100 random, wrong answers of \alpaca{} and \dvc{} on the ArchivalQA dataset concerning their failure types (see Figure~\ref{fig:manual_analysis_archivalqa}).

For nearly half of \alpaca{}'s wrong answers, we observe factually wrong yet plausible answers: These are usually questions where the model gives the \textbf{correct answer type but factually wrong entity} (such as a person or a location). As part of the plausible answers, we also find a set of \textit{temporally shifted} entities as answers, i.e.,  successor or predecessor of the correct entity. 
Next, we observe a set of \textbf{answers that represent uncertainty in the model} (``unknown'', ``I do not know'', etc.), followed by \textbf{technically correct answers with a different granularity than the ground truth} (e.g., answering with a country instead of a city when asked where a person was born). We also found a set of \textbf{implausible answers} where the model did not produce the right type of answer but repeated the question or produced incoherent and wrong text. Lastly, we also observed a few cases of \textbf{wrong ground truth answers}, which stem from the ArchivalQA dataset being automatically generated and cleaned. 
When inspecting the manual analysis results for \dvc{}, we observe many questions that were not answered - suggesting that \dvc{} is better calibrated not to answer in case of uncertainty. There are also fewer implausible answers and a similar amount of temporal errors.

\paragraph{\textbf{Insight.}} Temporal hallucinations occur when \alpaca{} and \dvc{} answer temporally-scoped questions. 

\subsubsection{Effect of Question Words}
\begin{figure}[ht]
    \begin{center}
        \scalebox{.89}{\input{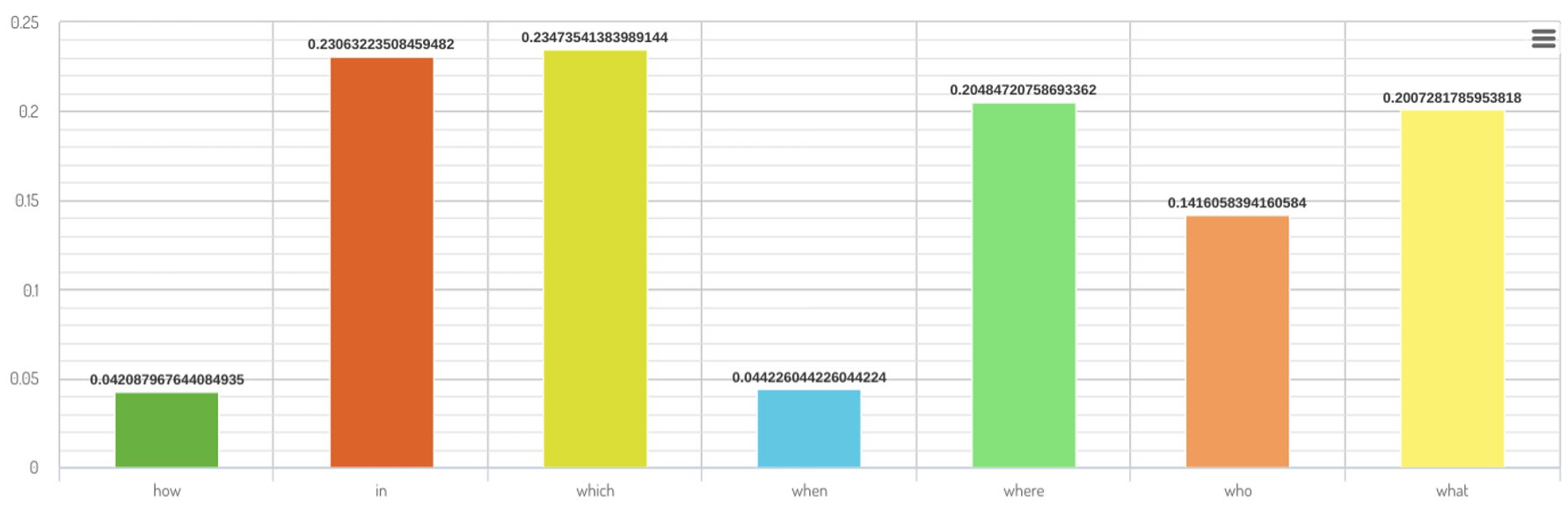}}
    \end{center}
    \caption{Performance of \alpaca{} and \dvc{} on ArchivalQA stratified by the leading words of questions.}
    \label{fig:question_words}
\end{figure}

Next, we analyze the performance of our model on different types of questions. To do so, we stratify the performance on ArchivalQA on the question's leading words (e.g., who, where, what) and present the results in Figure~\ref{fig:question_words}.
It is apparent that both models' performance on the question words "how" and "when" is much worse than for the other question words (i.e., "which," "where," "who," and "what"). We conjecture that "when" (very time-focused) and "how" (procedural or ambiguous) are more difficult than the other questions. 
Especially "when" requires the model to be aware of the time so as not to ignore the time references (both in questions and in training documents) compared to other question types for which it might be less critical, as we saw in our time reference randomization experiments (Figure~\ref{fig:absolute_relative_random_reference}). 
Interestingly, we find \dvc{} to perform similarly to the much smaller \alpaca{} on the temporal "when" questions. 

\paragraph{\textbf{Insight.}} \alpaca{} and \dvc{} perform poorly on questions requiring explicit temporal knowledge ("when").

\input{Tables/temporal_errors_examples}

\subsection{Types of Temporal Errors}

Next, we use our experiments and the manual error analysis to characterize the types of temporal errors exhibited by the models. Our analysis revealed four potential categories of temporal errors: \textit{temporal shifts, time invariance, temporal inertia, and referencing errors}. Table~\ref{tab:blind-spots} contains examples of these temporal errors.

\subsubsection{Temporal Shifts} Temporal shifts occur when the model incorrectly disambiguates the actual time of concern. For instance, when asked about the winners of the Oscars in 1994, the model might mistakenly predict the winners of 1995 instead. This error highlights the model's difficulty in accurately determining and aligning the specific temporal context with the question's intent.

\subsubsection{Time Invariance} Biases due to popularity arise from a strong semantic association between the answer entity and an entity mentioned in the question. This association often leads the model to disregard the temporal constraints specified in the question. 
In some cases, this bias can even override entity type constraints. 
For example, our analysis showed that the model incorrectly predicted Mike Tyson as the answer due to the strong association between boxing and Tyson, disregarding the temporal context.

\subsubsection{Temporal Inertia} Furthermore, we hypothesized that statistical support for certain past entity-entity relationships could cause LLMs to answer questions about the recent past incorrectly. We attributed these potential statistical errors to temporal inertia. This suggests that past statistical patterns might influence the model's predictions, resulting in inaccurate responses when the more recent past deviates from those patterns.

\subsubsection{Referencing Errors} 
Referencing errors occur due to an incorrect understanding of the current time.
Table~\ref{tab:blind-spots} shows an example of incorrect referencing of the past time.
Also, future time, defined as temporal intents after the model's training time, is beyond the scope of the model's parametric memory. Hence, the model is expected not to predict future events accurately.

\subsubsection{Measuring Temporal Errors}
We use the TempLAMA dataset to measure temporal errors since it contains ground truth for neighboring years, allowing us to test for temporal shifts. Specifically, we estimate temporal shifts by looking at examples where the object of the relation changes (e.g., a player changes his team) and whether the model predicts the next or previous object. Similarly, we investigate the extent of temporal inertia via the rate at which models fail to adapt to the last relation change of a subject. The amount of time-invariant relations is estimated by the ratio of relations for which the model always predicts the same output, no matter the year specified in the question. Lastly, we report the performance degradation of changing to relative references on the ArchivalQA dataset. The results are presented in Table~\ref{tab:temporal_errors}. 
\input{Tables/temporal_errors}
We observe that all categories of temporal errors frequently occur in our LLMs - posing new opportunities to improve the temporal abilities of LLMs. While temporal shifts seem rare, a relevant portion of facts is answered wrongly because the model did not incorporate new knowledge in its parametric memory (inertia). The most pronounced seems to be an invariance toward the time reference for many relations for all three models and the negative effects of using the relative time referencing. The latter is causing the performance to degrade by 11-35\%. 
While we offer a novel framework for temporal errors of LLMs and estimate their prevalence, a more fine-grained analysis of the causes will be necessary to alleviate these and to build capable and trustworthy language models. This, however, is out of the scope of this work. 

%% file: Figures/error_analysis/question_word_alpaca_archivalqa.tex
\begin{tikzpicture}

\definecolor{mint}{HTML}{29855E}
\definecolor{lightgreen}{HTML}{29855E}
\definecolor{purple}{HTML}{29855E}
\definecolor{organge}{HTML}{29855E}
\definecolor{sand}{HTML}{29855E}
\definecolor{steelblue31119180}{HTML}{29855E}
\definecolor{lightgray204}{RGB}{204,204,204}

\begin{axis}[
width=0.5\textwidth,
height=.2\textheight,
legend cell align={left},
legend style={
  fill opacity=0.8,
  draw opacity=1,
  text opacity=1,
  at={(0.03,0.97)},
  anchor=north west,
  draw=lightgray204
},
legend pos= south east,
tick align=outside,
tick pos=left,
x grid style={darkgray176},
xlabel={},
xmin=-0.5, xmax=5.5,
xtick style={color=black},
xtick={0,1,2,3,4,5},
xticklabels={how,which,when,where,who,what},
y grid style={darkgray176},
ylabel={Contains},
ymin=0, ymax=33,
ytick style={color=black},
]
\draw[draw=none,fill=lightgreen] (axis cs:-0.25,0) rectangle (axis cs:0,3.8);
\addlegendimage{ybar,ybar legend,draw=none,fill=absolute}
\addlegendentry{\alpaca{}}

\draw[draw=none,fill=mint] (axis cs:0.75,0) rectangle (axis cs:1,23.2);
\draw[draw=none,fill=purple] (axis cs:1.75,0) rectangle (axis cs:2,3.6);
\draw[draw=none,fill=organge] (axis cs:2.75,0) rectangle (axis cs:3,20.3);
\draw[draw=none,fill=sand] (axis cs:3.75,0) rectangle (axis cs:4,13.8);
\draw[draw=none,fill=steelblue31119180] (axis cs:4.75,0) rectangle (axis cs:5,20.2);

\draw[draw=none,fill=absolute_davinci] (axis cs:0,0) rectangle (axis cs:0.25,5.4);
\addlegendimage{ybar,ybar legend,draw=none,fill=absolute_davinci}
\addlegendentry{\dvc{}}
\draw[draw=none,fill=absolute_davinci] (axis cs:1,0) rectangle (axis cs:1.25,31.0);
\draw[draw=none,fill=absolute_davinci] (axis cs:2,0) rectangle (axis cs:2.25,3.8);
\draw[draw=none,fill=absolute_davinci] (axis cs:3,0) rectangle (axis cs:3.25,23.5);
\draw[draw=none,fill=absolute_davinci] (axis cs:4,0) rectangle (axis cs:4.25,20.7);
\draw[draw=none,fill=absolute_davinci] (axis cs:5,0) rectangle (axis cs:5.25,25.6);

\end{axis}

\end{tikzpicture}

%% file: Tables/temporal_errors_examples.tex
\begin{table*}[ht!]
\centering
 \small
\begin{tabular}{llll}
\toprule
\textbf{Model} & \textbf{Question} & \textbf{Temporal error} & \textbf{Explanation}\\ 
\toprule
            & \texttt{Tom Brady played for which team in 2020?}              &                        &\\
\alpaca{}   & \textcolor{magenta}{\texttt{New England Patriots}}                                  & Temporal inertia       & Brady joined Tampa (2020)  \\
Answer      & \textcolor{teal}{\texttt{Tampa Bay Buccaneers}}                                  &                        & after 20 years with Patriots\\
\midrule
            & \texttt{Cristiano Ronaldo played for which team in 2019?}                 &                       & \\
\rpbig{}      & \textcolor{magenta}{\texttt{Real Madrid}}                                   & Time invariance            & The model predicts Real \\ 
Answer      & \textcolor{teal}{\texttt{Juventus FC}}                                         &                       & Madrid for all years\\
\midrule
            & \makecell[l]{\texttt{In the 2003 Wimbledon Men's Singles} Tennis } &&\\

            &Championship, who beat Tim Henman?&Temporal shift&\\
\alpaca{}   & \textcolor{magenta}{\textsf{Andy Roddick}}                                          &  & Roddick beat Henman in\\
Answer      & 
\textcolor{teal}{\textsf{Sebastien Grosjean}}                                    &                        & the following year (2004)                 \\
\midrule
            & \makecell[l]{\texttt{Who painted a portrait of George} \texttt{Washington }}&                        &\\
            &for Martha in 1789?&Temporal shift&\\
\dvc{}      & \textcolor{magenta}{\texttt{Gilbert Stuart}}                                        &  & Stuart did paint Washington \\
Answer      & \textcolor{teal}{\texttt{John Ramage}}                                           &                        & but in 1795 and 1796\\ 
\midrule
            & \texttt{What state did Tuscany join 162 years ago?}	        &                        & \\
\dvc{}      & \textcolor{magenta}{\textsf{Unknown.}}                                             & Temporal referencing   & correct answer of Italy with\\
Answer      & \textcolor{teal}{\textsf{Italy}           }                                     &                        & absolute reference (1859)\\
\bottomrule
\end{tabular}
\caption{Examples of temporal blind spots.}
\label{tab:blind-spots}
\end{table*}

%% file: Tables/temporal_errors.tex
\begin{table}[ht]
\small
\begin{tabular}{lccc}
\hline
\textbf{Temp. Error}                & \alpaca{} & \dvc{} & \rpbig{} \\ \hline
Shift       & 5.0\%  & 6.3\%        & \textbf{3.9\%}       \\
Inertia     & 12.7\% & 18.8\%        & \textbf{9.2\%}       \\
Invariance      & \textbf{21.4\%} & 62.5\%        & 66.9\%       \\
Referencing & 30\%   & 35\%        & \textbf{11\%}       \\ \hline
\end{tabular}
\caption{Estimation of temporal error occurrences.}
\label{tab:temporal_errors}
\vspace{-4mm}
\end{table}

%% file: conclusion.tex
\section{Discussion and Conclusions}
\label{sec:conclusions}

In this study, we aimed to identify the degree of temporal understanding in Language Models (LLMs) by evaluating their performance on temporal QA tasks. Our findings shed light on several vital aspects of LLMs' temporal understanding abilities, offering insights into their limitations and blind spots.
First, we observed that LLMs exhibit lower effectiveness in answering questions about the past, particularly when detailed information is required. 
This indicates a relative weakness in their ability to recall accurately and reason about temporal aspects of historical events (cf. Table~\ref{tab:results_all}).
Furthermore, our investigation revealed that LLMs tend to capture more recent information better than older information up to a certain point. However, we hypothesize that the statistical support of older information can sometimes overshadow newer information, leading to a potential decline in performance (Figure~\ref{fig:time_strat}). 
To address this drawback, we propose that modeling the training data's creation and focus time could improve LLMs' temporal comprehension capabilities. 
We also find that relative time references proved more challenging for LLMs than absolute references. Task performance dropped by up to 35\% when relative time references were used, as demonstrated in Figure~\ref{fig:relative_absolute_time}. Moreover, randomizing the absolute time reference had a similar detrimental effect on model performance, resulting in a decrease of up to 55\%. By corrupting the time references, we showed that using relative, or absolute references that are wrong by more than a few years, results in worse model performance than using no time cues. 
In an extensive manual analysis, we find multiple temporal errors such as temporal shifts or the inability to update the parametric memory - and device automatic tests to detect them.
Our research underscores the importance of further improving LLMs' temporal understanding abilities.

%% file: appendix.tex
\section{Appendix}
\label{sec:appendix}

\input{Tables/templama_overview_shorter}

\input{Tables/prompts_overview}

\newpage

\begin{figure}
     \begin{subfigure}[b]{0.236\textwidth}
        \begin{center}
            \scalebox{.89}{\input{Figures/referencing/alternate/absolute_relative_random_reference_rpsmall}}
        \end{center}
        \caption{\rpsmall{}}
     \end{subfigure}
     \begin{subfigure}[b]{0.236\textwidth}
        \begin{center}
            \scalebox{.89}{\input{Figures/referencing/alternate/absolute_relative_random_reference_rpbig}}
        \end{center}
        \caption{\rpbig{}}
    \end{subfigure}
     \begin{subfigure}[b]{0.236\textwidth}
        \begin{center}
            \scalebox{.89}{\input{Figures/referencing/alternate/absolute_relative_random_reference_openllama}}
        \end{center}
        \caption{\openllama{}}
     \end{subfigure}
     \begin{subfigure}[b]{0.236\textwidth}
        \begin{center}
            \scalebox{.89}{\input{Figures/referencing/alternate/absolute_relative_random_reference_falcon}}
        \end{center}
        \caption{\falcon{}}
     \end{subfigure}
     \caption{Effect of randomized relative and absolute time references. Textured bars show the randomized variants.}
\end{figure}

\begin{figure*}
     \begin{subfigure}[b]{0.48\textwidth}
        \begin{center}
            \scalebox{.95}{\input{Figures/time_strat/archivalqa_line_em}}
        \end{center}
        \caption{ArchivalQA}
     \end{subfigure}
     \hfill
     \begin{subfigure}[b]{0.48\textwidth}
        \begin{center}
            \scalebox{.95}{\input{Figures/time_strat/templama_line_em}}
        \end{center}
        \caption{TempLAMA}
     \end{subfigure}
     \vskip\baselineskip
     \begin{subfigure}[b]{0.48\textwidth}
        \begin{center}
            \scalebox{.95}{\input{Figures/time_strat/archivalqa_line_f1}}
        \end{center}
        \caption{ArchivalQA}
     \end{subfigure}
     \hfill
     \begin{subfigure}[b]{0.48\textwidth}
        \begin{center}
            \scalebox{.95}{\input{Figures/time_strat/templama_line_f1}}
        \end{center}
        \caption{TempLAMA}
     \end{subfigure}
     \caption{Stratification of the \alpaca{}, \rpbig{}, and \dvc{} models on the ArchivalQA and TempLAMA datasets. Stratified by years, the trendline is the moving average with a window of 2. We do not show plots for the TemporalQuestions dataset since the dataset is not large enough for computing individual results per year.}
\end{figure*}

\begin{figure*}
     \begin{subfigure}[b]{0.48\textwidth}
        \begin{center}
            \scalebox{.95}{\input{Figures/time_strat/alternate/archivalqa_line_em}}
        \end{center}
        \caption{ArchivalQA}
     \end{subfigure}
     \hfill
     \begin{subfigure}[b]{0.48\textwidth}
        \begin{center}
            \scalebox{.95}{\input{Figures/time_strat/alternate/templama_line_em}}
        \end{center}
        \caption{TempLAMA}
     \end{subfigure}
     \vskip\baselineskip
     \begin{subfigure}[b]{0.48\textwidth}
        \begin{center}
            \scalebox{.95}{\input{Figures/time_strat/alternate/archivalqa_line_f1}}
        \end{center}
        \caption{ArchivalQA}
     \end{subfigure}
     \hfill
     \begin{subfigure}[b]{0.48\textwidth}
        \begin{center}
            \scalebox{.95}{\input{Figures/time_strat/alternate/templama_line_f1}}
        \end{center}
        \caption{TempLAMA}
     \end{subfigure}
     \vskip\baselineskip
     \begin{subfigure}[b]{0.48\textwidth}
        \begin{center}
            \scalebox{.95}{\input{Figures/time_strat/alternate/archivalqa_line_contains}}
        \end{center}
        \caption{ArchivalQA}
     \end{subfigure}
     \hfill
     \begin{subfigure}[b]{0.48\textwidth}
        \begin{center}
            \scalebox{.95}{\input{Figures/time_strat/alternate/templama_line_contains}}
        \end{center}
        \caption{TempLAMA}
     \end{subfigure}
     \caption{Stratification of the \falcon{}, \openllama{}, and \rpsmall{} models on the ArchivalQA and TempLAMA datasets. Stratified by years, the trendline is the moving average with a window of 2. We do not show plots for the TemporalQuestions dataset since the dataset is not large enough for computing individual results per year.}
\end{figure*}

%% file: Tables/templama_overview_shorter.tex
\begin{table}[ht]
\scriptsize
\begin{tabular}{lcl}
\hline
\textbf{\textbf{Relation}} & \textbf{\textbf{\#Qs}} & \textbf{\textbf{Template}}                                                                                    \\ \hline
member of sports team      & 9033                         & \textless{}sub.\textgreater~ played for which team in \textless{}year\textgreater{}?                 \\
position held              & 7343                         & \textless{}sub.\textgreater~ held which position in \textless{}year\textgreater{}?                         \\
employer                   & 9049                         & \textless{}sub.\textgreater~ worked for which company in \textless{}year\textgreater{}?                    \\
political party            & 7324                         & \textless{}sub.\textgreater~ was a member of which party in \textless{}year\textgreater{}?                 \\
head coach                 & 4886                         & \textless{}obj.\textgreater~ was the head coach of which team in \textless{}year\textgreater{}?             \\
educated at                & 1672                         & \textless{}sub.\textgreater~ attended which university in \textless{}year\textgreater{}?                   \\
chairperson                & 4190                         & \textless{}obj.\textgreater~ was the chair of which entity in \textless{}year\textgreater{}?                \\
head of government         & 4125                         & \makecell[l]{\textless{}obj.\textgreater~ is the head of the government of which state\\in \textless{}year\textgreater{}?} \\
owned by                   & 2688                         & \textless{}sub.\textgreater~ is owned by whom in \textless{}year\textgreater{}?                            \\ \hline
\end{tabular}
\caption{TempLAMA reformulations to natural questions.}
\label{}
\end{table}

%% file: Tables/prompts_overview.tex
\begin{table}[ht]
\Small
\begin{tabular}{l}
\toprule 
\makecell[l]{\textbf{\alpaca{}, \openllama{}}:
\\
\midrule
Below is an instruction that describes a task, paired with an input that 
provides \\further context. Write a response that appropriately 
completes the request.\\
\\
\#\#\# \textbf{Instruction:}\\ \texttt{Answer the question.}\\ \\ \#\#\# \textbf{Input:}\\ \{\textbf{davinci prompt} (except last line) -- see below\}\\ \\ \#\#\# \textbf{Response:}
} \\
\\
\midrule
\makecell[l]{\textbf{\dvc{}, \falcon{}, \rpsmall{}, \rpbig{}}: 
\\
\midrule
I am a highly intelligent question answering bot. If you ask me a question that is  \\
  rooted in truth, I will give you the answer. If you ask me a question that is \\  
 nonsense, trickery, or has no clear answer, I will respond with 'Unknown'.\\ \\ 
\texttt{Q: What is human life expectancy in the United} \texttt{States?}\\ A: 78 years\\ \\ \texttt{Q: Who was president of the United States in} \texttt{1955?}\\A: Dwight D. Eisenhower\\ \\ \texttt{Q: Which party did he belong to?}\\A: Republican Party\\ \\ \texttt{Q: Where were the 1992 Olympics held?}\\A: Barcelona, Spain.\\ \\Q: \{question\}\\A:
}\\
\bottomrule \\
\end{tabular}
\caption{Prompts used in this study. \{question\} is replaced with the current question.}
\end{table}

%% file: Figures/referencing/alternate/absolute_relative_random_reference_rpsmall.tex
\begin{tikzpicture}

\definecolor{darkgray176}{RGB}{176,176,176}
\definecolor{lightgray204}{RGB}{204,204,204}

\begin{axis}[
width=\textwidth,
height=.25\textheight,
legend cell align={left},
legend style={
  fill opacity=0.8,
  draw opacity=1,
  text opacity=1,
  at={(0.03,0.97)},
  anchor=north west,
  draw=lightgray204
},
tick align=outside,
tick pos=left,
x grid style={darkgray176},
xmin=-0.5, xmax=2.5,
xtick style={color=black},
xtick={0,1,2},
xticklabels={EM,F1,Contains},
y grid style={darkgray176},
ymin=0, ymax=20.5,
ytick style={color=black}
]
\draw[draw=none,fill=relative_redpajama_3,postaction={pattern=north east lines}] (axis cs:-0.25,0) rectangle (axis cs:-0.125,3.3);
\addlegendimage{ybar,ybar legend,draw=none,fill=relative_redpajama_3,postaction={pattern=north east lines}}
\addlegendentry{Rel.-rnd}

\draw[draw=none,fill=relative_redpajama_3,postaction={pattern=north east lines}] (axis cs:0.75,0) rectangle (axis cs:0.875,7.4);
\draw[draw=none,fill=relative_redpajama_3,postaction={pattern=north east lines}] (axis cs:1.75,0) rectangle (axis cs:1.875,5.7);
\draw[draw=none,fill=relative_redpajama_3] (axis cs:-0.125,0) rectangle (axis cs:0,3.76);
\addlegendimage{ybar,ybar legend,draw=none,fill=relative_redpajama_3}
\addlegendentry{Rel.}

\draw[draw=none,fill=relative_redpajama_3] (axis cs:0.875,0) rectangle (axis cs:1,7.7);
\draw[draw=none,fill=relative_redpajama_3] (axis cs:1.875,0) rectangle (axis cs:2,5.7);
\draw[draw=none,fill=absolute_redpajama_3,postaction={pattern=north east lines}] (axis cs:0,0) rectangle (axis cs:0.125,2.4);
\addlegendimage{ybar,ybar legend,draw=none,fill=absolute_redpajama_3,postaction={pattern=north east lines}}
\addlegendentry{Abs.-rnd}

\draw[draw=none,fill=absolute_redpajama_3,postaction={pattern=north east lines}] (axis cs:1,0) rectangle (axis cs:1.125,5.6);
\draw[draw=none,fill=absolute_redpajama_3,postaction={pattern=north east lines}] (axis cs:2,0) rectangle (axis cs:2.125,4.2);
\draw[draw=none,fill=absolute_redpajama_3] (axis cs:0.125,0) rectangle (axis cs:0.25,4);
\addlegendimage{ybar,ybar legend,draw=none,fill=absolute_redpajama_3}
\addlegendentry{Abs.}

\draw[draw=none,fill=absolute_redpajama_3] (axis cs:1.125,0) rectangle (axis cs:1.25,8.7);
\draw[draw=none,fill=absolute_redpajama_3] (axis cs:2.125,0) rectangle (axis cs:2.25,7.1);
\end{axis}

\end{tikzpicture}

%% file: Figures/referencing/alternate/absolute_relative_random_reference_rpbig.tex
\begin{tikzpicture}

\definecolor{darkgray176}{RGB}{176,176,176}
\definecolor{lightgray204}{RGB}{204,204,204}

\begin{axis}[
width=\textwidth,
height=.25\textheight,
legend cell align={left},
legend style={
  fill opacity=0.8,
  draw opacity=1,
  text opacity=1,
  at={(0.03,0.97)},
  anchor=north west,
  draw=lightgray204
},
tick align=outside,
tick pos=left,
x grid style={darkgray176},
xmin=-0.5, xmax=2.5,
xtick style={color=black},
xtick={0,1,2},
xticklabels={EM,F1,Contains},
y grid style={darkgray176},
ymin=0, ymax=24.5,
ytick style={color=black}
]
\draw[draw=none,fill=relative_redpajama_7,postaction={pattern=north east lines}] (axis cs:-0.25,0) rectangle (axis cs:-0.125,4.5);
\addlegendimage{ybar,ybar legend,draw=none,fill=relative_redpajama_7,postaction={pattern=north east lines}}
\addlegendentry{Rel.-rnd}

\draw[draw=none,fill=relative_redpajama_7,postaction={pattern=north east lines}] (axis cs:0.75,0) rectangle (axis cs:0.875,8.6);
\draw[draw=none,fill=relative_redpajama_7,postaction={pattern=north east lines}] (axis cs:1.75,0) rectangle (axis cs:1.875,7.2);
\draw[draw=none,fill=relative_redpajama_7] (axis cs:-0.125,0) rectangle (axis cs:0,5.9);
\addlegendimage{ybar,ybar legend,draw=none,fill=relative_redpajama_7}
\addlegendentry{Rel.}

\draw[draw=none,fill=relative_redpajama_7] (axis cs:0.875,0) rectangle (axis cs:1,10.8);
\draw[draw=none,fill=relative_redpajama_7] (axis cs:1.875,0) rectangle (axis cs:2,8.9);
\draw[draw=none,fill=absolute_redpajama_7,postaction={pattern=north east lines}] (axis cs:0,0) rectangle (axis cs:0.125,4.2);
\addlegendimage{ybar,ybar legend,draw=none,fill=absolute_redpajama_7,postaction={pattern=north east lines}}
\addlegendentry{Abs.-rnd}

\draw[draw=none,fill=absolute_redpajama_7,postaction={pattern=north east lines}] (axis cs:1,0) rectangle (axis cs:1.125,8.3);
\draw[draw=none,fill=absolute_redpajama_7,postaction={pattern=north east lines}] (axis cs:2,0) rectangle (axis cs:2.125,6.0);
\draw[draw=none,fill=absolute_redpajama_7] (axis cs:0.125,0) rectangle (axis cs:0.25,6.6);
\addlegendimage{ybar,ybar legend,draw=none,fill=absolute_redpajama_7}
\addlegendentry{Abs.}

\draw[draw=none,fill=absolute_redpajama_7] (axis cs:1.125,0) rectangle (axis cs:1.25,11.8);
\draw[draw=none,fill=absolute_redpajama_7] (axis cs:2.125,0) rectangle (axis cs:2.25,9.7);
\end{axis}

\end{tikzpicture}

%% file: Figures/referencing/alternate/absolute_relative_random_reference_openllama.tex
\begin{tikzpicture}

\definecolor{darkgray176}{RGB}{176,176,176}
\definecolor{lightgray204}{RGB}{204,204,204}

\begin{axis}[
width=\textwidth,
height=.25\textheight,
legend cell align={left},
legend style={
  fill opacity=0.8,
  draw opacity=1,
  text opacity=1,
  at={(0.03,0.97)},
  anchor=north west,
  draw=lightgray204
},
tick align=outside,
tick pos=left,
x grid style={darkgray176},
xmin=-0.5, xmax=2.5,
xtick style={color=black},
xtick={0,1,2},
xticklabels={EM,F1,Contains},
y grid style={darkgray176},
ymin=0, ymax=20.5,
ytick style={color=black}
]
\draw[draw=none,fill=relative_openllama,postaction={pattern=north east lines}] (axis cs:-0.25,0) rectangle (axis cs:-0.125,2.4);
\addlegendimage{ybar,ybar legend,draw=none,fill=relative_openllama,postaction={pattern=north east lines}}
\addlegendentry{Rel.-rnd}

\draw[draw=none,fill=relative_openllama,postaction={pattern=north east lines}] (axis cs:0.75,0) rectangle (axis cs:0.875,5.7);
\draw[draw=none,fill=relative_openllama,postaction={pattern=north east lines}] (axis cs:1.75,0) rectangle (axis cs:1.875,6.5);
\draw[draw=none,fill=relative_openllama] (axis cs:-0.125,0) rectangle (axis cs:0,3.1);
\addlegendimage{ybar,ybar legend,draw=none,fill=relative_openllama}
\addlegendentry{Rel.}

\draw[draw=none,fill=relative_openllama] (axis cs:0.875,0) rectangle (axis cs:1,6.8);
\draw[draw=none,fill=relative_openllama] (axis cs:1.875,0) rectangle (axis cs:2,7.7);
\draw[draw=none,fill=absolute_openllama,postaction={pattern=north east lines}] (axis cs:0,0) rectangle (axis cs:0.125,2.6);
\addlegendimage{ybar,ybar legend,draw=none,fill=absolute_openllama,postaction={pattern=north east lines}}
\addlegendentry{Abs.-rnd}

\draw[draw=none,fill=absolute_openllama,postaction={pattern=north east lines}] (axis cs:1,0) rectangle (axis cs:1.125,6.8);
\draw[draw=none,fill=absolute_openllama,postaction={pattern=north east lines}] (axis cs:2,0) rectangle (axis cs:2.125,7.2);
\draw[draw=none,fill=absolute_openllama] (axis cs:0.125,0) rectangle (axis cs:0.25,3.8);
\addlegendimage{ybar,ybar legend,draw=none,fill=absolute_openllama}
\addlegendentry{Abs.}

\draw[draw=none,fill=absolute_openllama] (axis cs:1.125,0) rectangle (axis cs:1.25,9.6);
\draw[draw=none,fill=absolute_openllama] (axis cs:2.125,0) rectangle (axis cs:2.25,10.9);
\end{axis}

\end{tikzpicture}

%% file: Figures/referencing/alternate/absolute_relative_random_reference_falcon.tex
\begin{tikzpicture}

\definecolor{darkgray176}{RGB}{176,176,176}
\definecolor{lightgray204}{RGB}{204,204,204}

\begin{axis}[
width=\textwidth,
height=.25\textheight,
legend cell align={left},
legend style={
  fill opacity=0.8,
  draw opacity=1,
  text opacity=1,
  at={(0.03,0.97)},
  anchor=north west,
  draw=lightgray204
},
tick align=outside,
tick pos=left,
x grid style={darkgray176},
xmin=-0.5, xmax=2.5,
xtick style={color=black},
xtick={0,1,2},
xticklabels={EM,F1,Contains},
y grid style={darkgray176},
ymin=0, ymax=20.5,
ytick style={color=black}
]
\draw[draw=none,fill=relative_falcon,postaction={pattern=north east lines}] (axis cs:-0.25,0) rectangle (axis cs:-0.125,1.6);
\addlegendimage{ybar,ybar legend,draw=none,fill=relative_falcon,postaction={pattern=north east lines}}
\addlegendentry{Rel.-rnd}

\draw[draw=none,fill=relative_falcon,postaction={pattern=north east lines}] (axis cs:0.75,0) rectangle (axis cs:0.875,7.7);
\draw[draw=none,fill=relative_falcon,postaction={pattern=north east lines}] (axis cs:1.75,0) rectangle (axis cs:1.875,6.5);
\draw[draw=none,fill=relative_falcon] (axis cs:-0.125,0) rectangle (axis cs:0,1.7);
\addlegendimage{ybar,ybar legend,draw=none,fill=relative_falcon}
\addlegendentry{Rel.}

\draw[draw=none,fill=relative_falcon] (axis cs:0.875,0) rectangle (axis cs:1,8.0);
\draw[draw=none,fill=relative_falcon] (axis cs:1.875,0) rectangle (axis cs:2,6.9);
\draw[draw=none,fill=absolute_falcon,postaction={pattern=north east lines}] (axis cs:0,0) rectangle (axis cs:0.125,2.2);
\addlegendimage{ybar,ybar legend,draw=none,fill=absolute_falcon,postaction={pattern=north east lines}}
\addlegendentry{Abs.-rnd}

\draw[draw=none,fill=absolute_falcon,postaction={pattern=north east lines}] (axis cs:1,0) rectangle (axis cs:1.125,8.0);
\draw[draw=none,fill=absolute_falcon,postaction={pattern=north east lines}] (axis cs:2,0) rectangle (axis cs:2.125,5.7);
\draw[draw=none,fill=absolute_falcon] (axis cs:0.125,0) rectangle (axis cs:0.25,2.9);
\addlegendimage{ybar,ybar legend,draw=none,fill=absolute_falcon}
\addlegendentry{Abs.}

\draw[draw=none,fill=absolute_falcon] (axis cs:1.125,0) rectangle (axis cs:1.25,10.1);
\draw[draw=none,fill=absolute_falcon] (axis cs:2.125,0) rectangle (axis cs:2.25,8.2);

\end{axis}
\end{tikzpicture}

%% file: Figures/time_strat/archivalqa_line_em.tex
\begin{tikzpicture}

    \definecolor{darkgray176}{RGB}{176,176,176}
    \definecolor{steelblue31119180}{RGB}{31,119,180}
    \definecolor{lightgray204}{RGB}{204,204,204}
    
    \begin{axis}[
    width=.99\textwidth,
    height=.2\textheight,
    legend cell align={left},
    legend style={
      fill opacity=0.8,
      draw opacity=1,
      text opacity=1,
      at={(0.03,0.97)},
      anchor=north west,
      draw=lightgray204
    },
    legend pos= south east,
    tick align=outside,
    tick pos=left,
    x grid style={darkgray176},
    xlabel={},
    xmin=-0.5, xmax=20.5,
    xtick style={color=black},
    xtick={
        0,
        2,
        4,
        6,
        8,
        10,
        12,
        14,
        16,
        18,
        20
    },
    xticklabel style={rotate=90.0},
    xticklabels={
      {1987},
      {1989},
      {1991},
      {1993},
      {1995},
      {1997},
      {1999},
      {2001},
      {2003},
      {2005},
      {2007}
    },
    y grid style={darkgray176},
    ylabel={EM},
    ymin=0, ymax=13,
    ytick style={color=black}
    ]
    
\addplot [thick, absolute_alpaca, mark=diamond]
table {
0          NaN
1     3.095098
2     3.218836
3     4.034401
4     5.579118
5     5.194472
6     3.472590
7     3.747725
8     4.189167
9     3.249501
10    3.226457
11    4.555890
12    5.670186
13    4.899360
14    4.248068
15    4.674155
16    5.028846
17    5.222568
18    4.928054
19    5.922069
20    6.470588
};
\addlegendentry{{\scriptsize \alpaca{}}}
\addplot [thick, absolute_davinci]
table {
0           NaN
1      7.699648
2      9.101040
3     10.096747
4     11.561230
5     10.251764
6      7.348387
7      8.629556
8      9.271401
9      7.719976
10     8.725924
11     9.962537
12    10.251812
13     9.747230
14     8.763311
15     9.197653
16     9.395975
17     9.610754
18     9.465133
19    10.839572
20    11.294118
};
\addlegendentry{{\scriptsize \dvc{}}}
\addplot [thick, absolute_redpajama_7, mark=triangle] 
table {
0          NaN
1     3.772147
2     4.653344
3     5.391977
4     5.982149
5     5.575059
6     4.244795
7     4.390159
8     4.989551
9     4.806566
10    5.790325
11    6.498736
12    6.428952
13    6.268139
14    5.871744
15    5.992436
16    6.054730
17    6.283410
18    7.126087
19    8.485249
20    7.505882
};
\addlegendentry{{\scriptsize \rpbig{}}}
    
\end{axis}

\end{tikzpicture}

%% file: Figures/time_strat/templama_line_em.tex
\begin{tikzpicture}

\definecolor{darkgray176}{RGB}{176,176,176}
\definecolor{steelblue31119180}{RGB}{31,119,180}
\definecolor{lightgray204}{RGB}{204,204,204}

\begin{axis}[
width=.99\textwidth,
height=.2\textheight,
legend cell align={left},
legend style={
  fill opacity=0.8,
  draw opacity=1,
  text opacity=1,
  at={(0.03,0.97)},
  anchor=north west,
  draw=lightgray204
},
legend pos= south west,
tick align=outside,
tick pos=left,
x grid style={darkgray176},
xlabel={},
xmin=-0.5, xmax=10.5,
xtick style={color=black},
xtick={0,1,2,3,4,5,6,7,8,9,10},
xticklabel style={rotate=90.0},
xticklabels={
  {2010},
  {2011},
  {2012},
  {2013},
  {2014},
  {2015},
  {2016},
  {2017},
  {2018},
  {2019},
  {2020}
},
y grid style={darkgray176},
ylabel={EM},
ymin=0, ymax=20,
ytick style={color=black}
]

\addplot [thick, absolute_alpaca, mark=diamond]
table {
0          NaN
1     3.469181
2     3.525070
3     3.590032
4     3.811985
5     4.091871
6     4.262701
7     4.402322
8     4.249395
9     3.914092
10    3.386714
};
\addlegendentry{{\scriptsize \alpaca{}}}
\addplot [thick, absolute_davinci]
table {
0           NaN
1     15.873044
2     16.099917
3     16.467239
4     17.009069
5     17.274918
6     16.956098
7     16.675520
8     16.372297
9     15.948167
10    15.233147
};
\addlegendentry{{\scriptsize \dvc{}}}
\addplot [thick, absolute_redpajama_7, mark=triangle] 
table {
0           NaN
1     10.665504
2     10.765747
3     10.994836
4     11.457680
5     11.811769
6     11.789727
7     11.804932
8     11.412594
9     10.555065
10     9.872787
};
\addlegendentry{{\scriptsize \rpbig{}}}
\end{axis}

\end{tikzpicture}

%% file: Figures/time_strat/archivalqa_line_f1.tex
\begin{tikzpicture}

    \definecolor{darkgray176}{RGB}{176,176,176}
    \definecolor{steelblue31119180}{RGB}{31,119,180}
    \definecolor{lightgray204}{RGB}{204,204,204}
    
    \begin{axis}[
    width=.99\textwidth,
    height=.2\textheight,
    legend cell align={left},
    legend style={
      fill opacity=0.8,
      draw opacity=1,
      text opacity=1,
      at={(0.03,0.97)},
      anchor=north west,
      draw=lightgray204
    },
    legend pos= south east,
    tick align=outside,
    tick pos=left,
    x grid style={darkgray176},
    xlabel={},
    xmin=-0.5, xmax=20.5,
    xtick style={color=black},
    xtick={
        0,
        2,
        4,
        6,
        8,
        10,
        12,
        14,
        16,
        18,
        20
    },
    xticklabel style={rotate=90.0},
    xticklabels={
      {1987},
      {1989},
      {1991},
      {1993},
      {1995},
      {1997},
      {1999},
      {2001},
      {2003},
      {2005},
      {2007}
    },
    y grid style={darkgray176},
    ylabel={F1},
    ymin=0, ymax=22,
    ytick style={color=black}
    ]
    
\addplot [thick, absolute_alpaca, mark=diamond] 
table {
0           NaN
1      9.693879
2      9.832880
3     10.450222
4     12.812311
5     12.192602
6      9.513230
7     10.042301
8     10.644029
9      9.949549
10    10.033557
11    11.378848
12    12.380656
13    11.479613
14    11.070645
15    11.545026
16    11.837908
17    11.989097
18    11.749982
19    11.974907
20    12.360651
};
\addlegendentry{{\scriptsize \alpaca{}}}
\addplot [thick, absolute_davinci] 
table {
0           NaN
1     16.681298
2     18.226218
3     19.222936
4     21.445293
5     19.853926
6     16.072781
7     18.099840
8     18.778948
9     16.899769
10    17.552923
11    19.067034
12    19.833037
13    19.659069
14    18.630323
15    18.875985
16    19.418439
17    19.208797
18    18.714016
19    20.145381
20    20.183554
};
\addlegendentry{{\scriptsize \dvc{}}}
\addplot [thick, absolute_redpajama_7, mark=triangle] 
table {
0           NaN
1      8.513226
2      9.199474
3      9.145537
4     10.205391
5     10.012520
6      8.195775
7      8.436159
8      9.082175
9      9.121965
10    10.296565
11    10.931413
12    10.872710
13    10.708713
14    10.791537
15    10.731631
16    11.204984
17    11.715647
18    11.966957
19    13.215966
20    12.938095
};
\addlegendentry{{\scriptsize \rpbig{}}}
    
\end{axis}

\end{tikzpicture}

%% file: Figures/time_strat/templama_line_f1.tex
\begin{tikzpicture}

\definecolor{darkgray176}{RGB}{176,176,176}
\definecolor{steelblue31119180}{RGB}{31,119,180}
\definecolor{lightgray204}{RGB}{204,204,204}

\begin{axis}[
width=.99\textwidth,
height=.2\textheight,
legend cell align={left},
legend style={
  fill opacity=0.8,
  draw opacity=1,
  text opacity=1,
  at={(0.03,0.97)},
  anchor=north west,
  draw=lightgray204
},
legend pos= south west,
tick align=outside,
tick pos=left,
x grid style={darkgray176},
xlabel={},
xmin=-0.5, xmax=10.5,
xtick style={color=black},
xtick={0,1,2,3,4,5,6,7,8,9,10},
xticklabel style={rotate=90.0},
xticklabels={
  {2010},
  {2011},
  {2012},
  {2013},
  {2014},
  {2015},
  {2016},
  {2017},
  {2018},
  {2019},
  {2020}
},
y grid style={darkgray176},
ylabel={F1},
ymin=0, ymax=34,
ytick style={color=black}
]

\addplot [thick, absolute_alpaca, mark=diamond] 
table {
0           NaN
1     16.934682
2     16.947185
3     16.821597
4     17.044148
5     17.437598
6     17.419229
7     17.241148
8     16.441350
9     15.379093
10    14.123755
};
\addlegendentry{{\scriptsize \alpaca{}}}
\addplot [thick, absolute_davinci] 
table {
0           NaN
1     31.056156
2     31.291927
3     31.642849
4     32.110113
5     32.094371
6     31.739515
7     31.294814
8     30.460699
9     29.429643
10    27.969857
};
\addlegendentry{{\scriptsize \dvc{}}}
\addplot [thick, absolute_redpajama_7, mark=triangle] 
table {
0           NaN
1     19.565255
2     19.467492
3     19.418291
4     19.582176
5     19.980414
6     19.944975
7     19.775739
8     18.954836
9     17.480047
10    16.614257
};
\addlegendentry{{\scriptsize \rpbig{}}}
\end{axis}

\end{tikzpicture}

%% file: Figures/time_strat/alternate/archivalqa_line_em.tex
\begin{tikzpicture}

    \definecolor{darkgray176}{RGB}{176,176,176}
    \definecolor{steelblue31119180}{RGB}{31,119,180}
    \definecolor{lightgray204}{RGB}{204,204,204}
    
    \begin{axis}[
    width=.99\textwidth,
    height=.2\textheight,
    legend cell align={left},
    legend style={
      fill opacity=0.8,
      draw opacity=1,
      text opacity=1,
      at={(0.03,0.97)},
      anchor=north west,
      draw=lightgray204
    },
    legend pos= south east,
    tick align=outside,
    tick pos=left,
    x grid style={darkgray176},
    xlabel={},
    xmin=-0.5, xmax=20.5,
    xtick style={color=black},
    xtick={
        0,
        2,
        4,
        6,
        8,
        10,
        12,
        14,
        16,
        18,
        20
    },
    xticklabel style={rotate=90.0},
    xticklabels={
      {1987},
      {1989},
      {1991},
      {1993},
      {1995},
      {1997},
      {1999},
      {2001},
      {2003},
      {2005},
      {2007}
    },
    y grid style={darkgray176},
    ylabel={EM},
    ymin=0, ymax=8,
    ytick style={color=black}
    ]
\addplot [thick, absolute_falcon, mark=o] 
table {
0          NaN
1     1.399946
2     1.796070
3     2.594026
4     2.935228
5     2.332516
6     1.898110
7     2.303653
8     2.656096
9     2.155311
10    2.249797
11    2.726071
12    2.860448
13    2.740287
14    2.715373
15    2.885882
16    2.737140
17    3.079825
18    3.187930
19    3.128162
20    3.247059
};
\addlegendentry{{\scriptsize \falcon{}}}
\addplot [thick, absolute_openllama, mark=square] 
table {
0          NaN
1     1.861500
2     2.760236
3     3.167375
4     3.176336
5     3.128252
6     2.215977
7     2.703624
8     3.432662
9     2.844027
10    2.927707
11    2.715956
12    2.556996
13    2.982525
14    3.178603
15    3.559762
16    3.075399
17    3.089934
18    3.439806
19    3.943940
20    4.611765
};
\addlegendentry{{\scriptsize \openllama{}}}
\addplot [thick, absolute_redpajama_3] 
table {
0          NaN
1     2.319317
2     2.691819
3     3.153153
4     3.476371
5     3.244847
6     2.753364
7     3.277992
8     3.788975
9     3.193330
10    3.345864
11    3.998159
12    3.719737
13    3.240755
14    3.743641
15    4.491974
16    4.158895
17    3.767090
18    3.762103
19    5.081656
20    6.694118
};
\addlegendentry{{\scriptsize \rpsmall{}}}
    
\end{axis}

\end{tikzpicture}

%% file: Figures/time_strat/alternate/templama_line_em.tex
\begin{tikzpicture}

\definecolor{darkgray176}{RGB}{176,176,176}
\definecolor{steelblue31119180}{RGB}{31,119,180}
\definecolor{lightgray204}{RGB}{204,204,204}

\begin{axis}[
width=.99\textwidth,
height=.2\textheight,
legend cell align={left},
legend style={
  fill opacity=0.8,
  draw opacity=1,
  text opacity=1,
  at={(0.03,0.97)},
  anchor=north west,
  draw=lightgray204
},
legend pos= south west,
tick align=outside,
tick pos=left,
x grid style={darkgray176},
xlabel={},
xmin=-0.5, xmax=10.5,
xtick style={color=black},
xtick={0,1,2,3,4,5,6,7,8,9,10},
xticklabel style={rotate=90.0},
xticklabels={
  {2010},
  {2011},
  {2012},
  {2013},
  {2014},
  {2015},
  {2016},
  {2017},
  {2018},
  {2019},
  {2020}
},
y grid style={darkgray176},
ylabel={EM},
ymin=0, ymax=6,
ytick style={color=black}
]
\addplot [thick, absolute_falcon, mark=o] 
table {
0          NaN
1     2.966165
2     3.169226
3     3.203508
4     3.150318
5     3.077119
6     3.038778
7     3.105044
8     3.000729
9     2.831571
10    2.799836
};
\addlegendentry{{\scriptsize \falcon{}}}
\addplot [thick, absolute_openllama, mark=square] 
table {
0          NaN
1     3.840623
2     3.698916
3     3.547023
4     3.649213
5     3.865166
6     4.058433
7     4.062954
8     3.766595
9     3.372589
10    3.085049
};
\addlegendentry{{\scriptsize \openllama{}}}
\addplot [thick, absolute_redpajama_3]
table {
0          NaN
1     3.955278
2     3.917777
3     3.636384
4     3.876369
5     4.415682
6     5.002179
7     5.497479
8     5.475574
9     5.267486
10    4.754615
};
\addlegendentry{{\scriptsize \rpsmall{}}}
\end{axis}

\end{tikzpicture}

%% file: Figures/time_strat/alternate/archivalqa_line_f1.tex
\begin{tikzpicture}

    \definecolor{darkgray176}{RGB}{176,176,176}
    \definecolor{steelblue31119180}{RGB}{31,119,180}
    \definecolor{lightgray204}{RGB}{204,204,204}
    
    \begin{axis}[
    width=.99\textwidth,
    height=.2\textheight,
    legend cell align={left},
    legend style={
      fill opacity=0.8,
      draw opacity=1,
      text opacity=1,
      at={(0.03,0.97)},
      anchor=north west,
      draw=lightgray204
    },
    legend pos= south east,
    tick align=outside,
    tick pos=left,
    x grid style={darkgray176},
    xlabel={},
    xmin=-0.5, xmax=20.5,
    xtick style={color=black},
    xtick={
        0,
        2,
        4,
        6,
        8,
        10,
        12,
        14,
        16,
        18,
        20
    },
    xticklabel style={rotate=90.0},
    xticklabels={
      {1987},
      {1989},
      {1991},
      {1993},
      {1995},
      {1997},
      {1999},
      {2001},
      {2003},
      {2005},
      {2007}
    },
    y grid style={darkgray176},
    ylabel={F1},
    ymin=0, ymax=11,
    ytick style={color=black}
    ]
\addplot [thick, absolute_falcon, mark=o]
table {
0          NaN
1     7.092923
2     7.699282
3     8.285602
4     8.633619
5     8.056019
6     7.014268
7     7.615136
8     8.559998
9     7.870091
10    7.724638
11    8.329665
12    8.331914
13    8.458135
14    8.493404
15    8.670458
16    8.723441
17    9.245018
18    9.434186
19    8.524952
20    7.837921
};
\addlegendentry{{\scriptsize \falcon{}}}
\addplot [thick, absolute_openllama, mark=square] 
table {
0          NaN
1     6.107622
2     7.032399
3     7.271892
4     8.059175
5     8.168032
6     6.417824
7     6.725923
8     7.819650
9     7.431619
10    7.287227
11    7.003036
12    6.845620
13    7.547158
14    8.080440
15    7.963394
16    7.463732
17    7.683479
18    7.608575
19    7.838174
20    8.337480
};
\addlegendentry{{\scriptsize \openllama{}}}
\addplot [thick, absolute_redpajama_3]
table {
0           NaN
1      6.509418
2      6.911242
3      7.491846
4      8.088351
5      8.011585
6      7.238939
7      7.450143
8      8.049881
9      7.709402
10     7.730566
11     8.134579
12     7.735535
13     7.683209
14     8.430913
15     8.460355
16     8.510374
17     8.812155
18     8.448523
19     9.184481
20    10.918658
};
\addlegendentry{{\scriptsize \rpsmall{}}}
    
\end{axis}

\end{tikzpicture}

%% file: Figures/time_strat/alternate/templama_line_f1.tex
\begin{tikzpicture}

\definecolor{darkgray176}{RGB}{176,176,176}
\definecolor{steelblue31119180}{RGB}{31,119,180}
\definecolor{lightgray204}{RGB}{204,204,204}

\begin{axis}[
width=.99\textwidth,
height=.2\textheight,
legend cell align={left},
legend style={
  fill opacity=0.8,
  draw opacity=1,
  text opacity=1,
  at={(0.03,0.97)},
  anchor=north west,
  draw=lightgray204
},
legend pos= south west,
tick align=outside,
tick pos=left,
x grid style={darkgray176},
xlabel={},
xmin=-0.5, xmax=10.5,
xtick style={color=black},
xtick={0,1,2,3,4,5,6,7,8,9,10},
xticklabel style={rotate=90.0},
xticklabels={
  {2010},
  {2011},
  {2012},
  {2013},
  {2014},
  {2015},
  {2016},
  {2017},
  {2018},
  {2019},
  {2020}
},
y grid style={darkgray176},
ylabel={F1},
ymin=0, ymax=15,
ytick style={color=black}
]
\addplot [thick, absolute_falcon, mark=o]
table {
0           NaN
1     13.431681
2     13.450226
3     13.295138
4     13.447313
5     13.380454
6     13.099757
7     13.238854
8     12.982437
9     12.096777
10    11.206534
};
\addlegendentry{{\scriptsize \falcon{}}}
\addplot [thick, absolute_openllama, mark=square] 
table {
0           NaN
1     13.721442
2     13.415806
3     13.233372
4     13.324204
5     13.324946
6     13.271166
7     13.219122
8     12.718735
9     12.021282
10    11.358684
};
\addlegendentry{{\scriptsize \openllama{}}}
\addplot [thick, absolute_redpajama_3] 
table {
0           NaN
1     10.332674
2      9.988580
3      9.163909
4      9.680834
5     10.803945
6     11.907085
7     12.699487
8     12.369742
9     11.670489
10    10.488555
};
\addlegendentry{{\scriptsize \rpsmall{}}}
\end{axis}

\end{tikzpicture}

%% file: Figures/time_strat/alternate/archivalqa_line_contains.tex
\begin{tikzpicture}

    \definecolor{darkgray176}{RGB}{176,176,176}
    \definecolor{steelblue31119180}{RGB}{31,119,180}
    \definecolor{lightgray204}{RGB}{204,204,204}
    
    \begin{axis}[
    width=.99\textwidth,
    height=.2\textheight,
    legend cell align={left},
    legend style={
      fill opacity=0.8,
      draw opacity=1,
      text opacity=1,
      at={(0.03,0.97)},
      anchor=north west,
      draw=lightgray204
    },
    legend pos= south east,
    tick align=outside,
    tick pos=left,
    x grid style={darkgray176},
    xlabel={},
    xmin=-0.5, xmax=20.5,
    xtick style={color=black},
    xtick={
        0,
        2,
        4,
        6,
        8,
        10,
        12,
        14,
        16,
        18,
        20
    },
    xticklabel style={rotate=90.0},
    xticklabels={
      {1987},
      {1989},
      {1991},
      {1993},
      {1995},
      {1997},
      {1999},
      {2001},
      {2003},
      {2005},
      {2007}
    },
    y grid style={darkgray176},
    ylabel={Contains},
    ymin=0, ymax=12,
    ytick style={color=black}
    ]
\addplot [thick, absolute_falcon, mark=o]
table {
0          NaN
1     5.838037
2     6.605498
3     6.461621
4     6.508578
5     6.099891
6     5.533549
7     6.381599
8     7.007524
9     6.093527
10    6.371437
11    7.211201
12    6.960420
13    7.211343
14    7.358500
15    7.346331
16    7.374299
17    8.050202
18    7.721523
19    7.194316
20    8.023529
};
\addlegendentry{{\scriptsize \falcon{}}}
\addplot [thick, absolute_openllama, mark=square]
table {
0           NaN
1      7.014903
2      7.966677
3      7.899678
4      8.378987
5      9.032512
6      7.454122
7      8.088379
8      9.254744
9      8.488039
10     8.150566
11     7.937568
12     7.973004
13     8.564909
14     9.410295
15     9.548917
16     9.426873
17    10.193712
18     9.371876
19     8.807186
20    10.623529
};
\addlegendentry{{\scriptsize \openllama{}}}
\addplot [thick, absolute_redpajama_3] 
table {
0          NaN
1     4.922295
2     5.601139
3     5.957760
4     6.629187
5     6.895474
6     6.300212
7     6.354033
8     6.978779
9     6.670984
10    6.727986
11    6.838548
12    6.631164
13    7.036199
14    7.194159
15    7.045420
16    7.862862
17    8.555350
18    7.450298
19    7.519438
20    9.552941
};
\addlegendentry{{\scriptsize \rpsmall{}}}
\addlegendentry{{\scriptsize \rpbig{}}}
    
\end{axis}

\end{tikzpicture}

%% file: Figures/time_strat/alternate/templama_line_contains.tex
\begin{tikzpicture}

\definecolor{darkgray176}{RGB}{176,176,176}
\definecolor{steelblue31119180}{RGB}{31,119,180}
\definecolor{lightgray204}{RGB}{204,204,204}

\begin{axis}[
width=.99\textwidth,
height=.2\textheight,
legend cell align={left},
legend style={
  fill opacity=0.8,
  draw opacity=1,
  text opacity=1,
  at={(0.03,0.97)},
  anchor=north west,
  draw=lightgray204
},
legend pos= south west,
tick align=outside,
tick pos=left,
x grid style={darkgray176},
xlabel={},
xmin=-0.5, xmax=10.5,
xtick style={color=black},
xtick={0,1,2,3,4,5,6,7,8,9,10},
xticklabel style={rotate=90.0},
xticklabels={
  {2010},
  {2011},
  {2012},
  {2013},
  {2014},
  {2015},
  {2016},
  {2017},
  {2018},
  {2019},
  {2020}
},
y grid style={darkgray176},
ylabel={Contains},
ymin=0, ymax=10,
ytick style={color=black}
]
\addplot [thick, absolute_falcon, mark=o] 
table {
0          NaN
1     6.065665
2     6.181677
3     6.259715
4     6.463069
5     6.672262
6     6.861374
7     7.282805
8     7.208742
9     6.558142
10    6.212909
};
\addlegendentry{{\scriptsize \falcon{}}}
\addplot [thick, absolute_openllama, mark=square]
table {
0          NaN
1     7.056537
2     7.097869
3     7.004351
4     7.125415
5     7.104377
6     7.011705
7     7.220057
8     7.185498
9     6.860159
10    6.368221
};
\addlegendentry{{\scriptsize \openllama{}}}
\addplot [thick, absolute_redpajama_3] 
table {
0          NaN
1     6.087188
2     6.206124
3     6.018485
4     6.406886
5     6.931423
6     7.396782
7     8.017217
8     8.128799
9     7.890819
10    7.428955
};
\addlegendentry{{\scriptsize \rpsmall{}}}
\end{axis}

\end{tikzpicture}